\documentclass{article}

\PassOptionsToPackage{numbers, compress}{natbib}

\usepackage[preprint]{neurips_2025}

\usepackage[utf8]{inputenc} 
\usepackage[T1]{fontenc}    
\usepackage{hyperref}       
\usepackage{url}            
\usepackage{booktabs}       
\usepackage{amsfonts}       
\usepackage{nicefrac}       
\usepackage{microtype}      
\usepackage{xcolor}         

\usepackage{multicol,multirow}
\usepackage{listings}
\usepackage{xcolor}
\usepackage{caption}
\usepackage{subcaption}
\usepackage{amsmath}
\usepackage{adjustbox}
\usepackage{wrapfig}
\usepackage{colortbl}
\usepackage{geometry}

\definecolor{purple1}{rgb}{0.8, 0.7, 1}
\definecolor{purple2}{rgb}{0.85, 0.75, 1}
\definecolor{purple3}{rgb}{0.9, 0.8, 1}
\definecolor{normalblue}{RGB}{0,102,204}
\definecolor{abnormalgreen}{RGB}{0,153,0}
\definecolor{browncolor}{RGB}{139,69,19}
\title{SurveillanceVQA-589K: A Benchmark for Comprehensive Surveillance Video-Language Understanding with Large Models}

\author{%
  Bo Liu$^1$, Pengfei Qiao$^1$, Minhan Ma$^1$, Xuange Zhang$^1$, Yinan Tang$^2$, \\ \textbf{Peng Xu$^3$, Kun Liu$^4$, Tongtong Yuan$^1$}\thanks{Corresponding author, Beijing University of Technology, E-mail: yuantt@bjut.edu.cn } \\
  $^1$ Department of Computer Science, Beijing University of Technology\\
  $^2$ Inspur Electronic Information Industry Co., Ltd \\
  $^3$ Department of Electronic Engineering, Tsinghua University\\
  $^4$ JD Explore Academy \\
}

\begin{document}

\maketitle
\begin{abstract}
Understanding surveillance video content remains a critical yet underexplored challenge in vision–language research, particularly due to its real-world complexity, irregular event dynamics, and safety-critical implications. In this work, we introduce SurveillanceVQA-589K, the largest open-ended video question answering (VQA) benchmark tailored to the surveillance domain. The dataset comprises 589,380 QA pairs spanning 12 cognitively diverse question types, including temporal reasoning, causal inference, spatial understanding, and anomaly interpretation, across both normal and abnormal video scenarios. To construct the benchmark at scale, we design a hybrid annotation pipeline that combines temporally aligned human-written captions with Large Vision-Language Model~(LVLM) assisted QA generation using prompt-based techniques. We also propose a multi-dimensional evaluation protocol to assess contextual, temporal, and causal comprehension. We evaluate eight LVLMs under this framework, revealing significant performance gaps, especially in causal and anomaly-related tasks, underscoring the limitations of current models in real-world surveillance contexts. Our benchmark provides a practical and comprehensive resource for advancing video-language understanding in safety-critical applications such as intelligent monitoring, incident analysis, and autonomous decision-making. The dataset and code are publicly available at: \url{https://huggingface.co/datasets/fei213/SurveillanceVQA-589K}.
\end{abstract}

\begin{figure}[htbp]
    \centering
    \begin{subfigure}[b]{0.48\textwidth}
        \centering
        \includegraphics[width=0.9\textwidth]{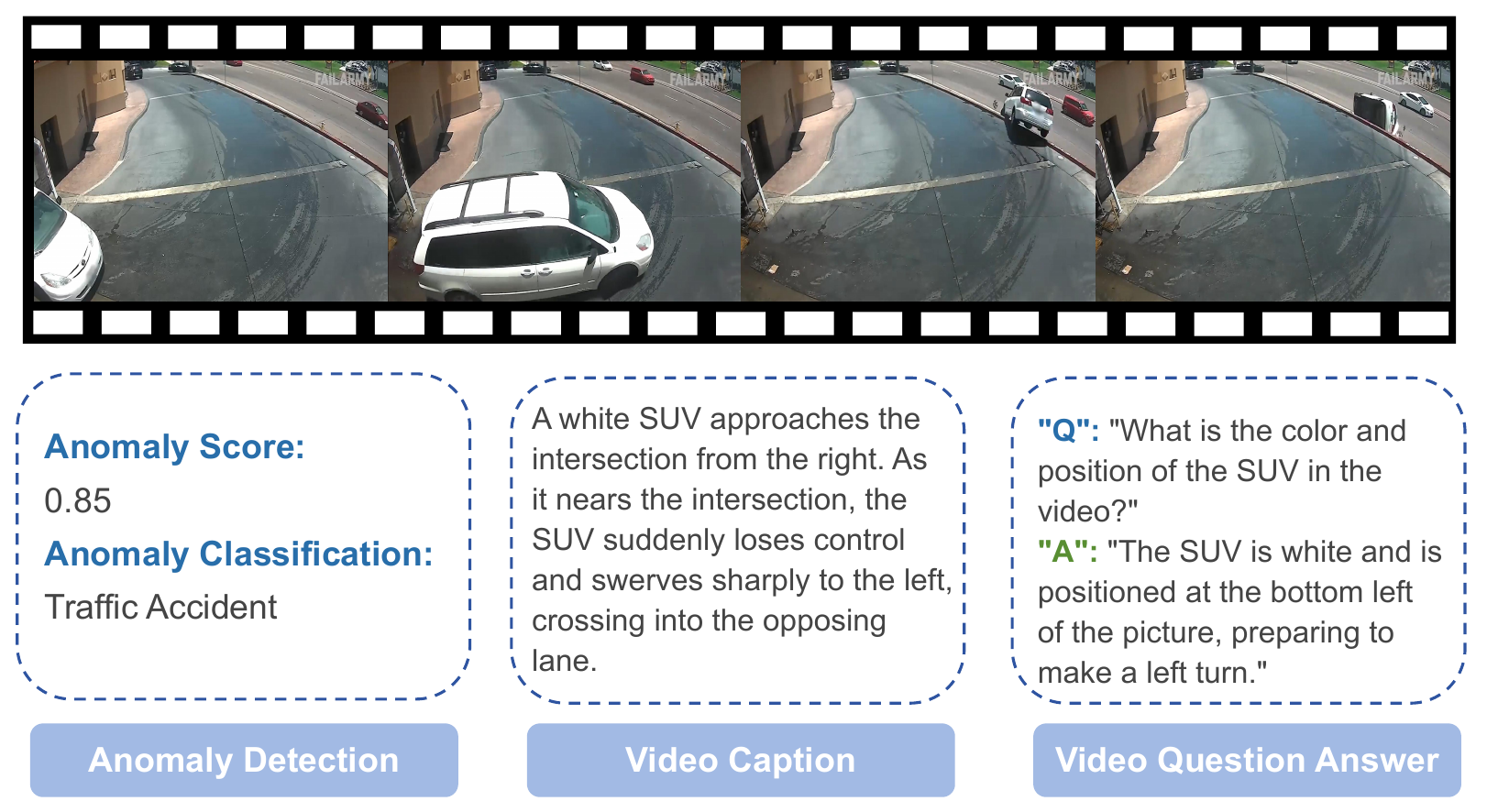}
        \caption{Anomaly detection, video caption vs. our VQA.}
        \label{fig:sub-a}
    \end{subfigure}
    \hfill
    \begin{subfigure}[b]{0.48\textwidth}
        \centering       
       \includegraphics[width=\textwidth]{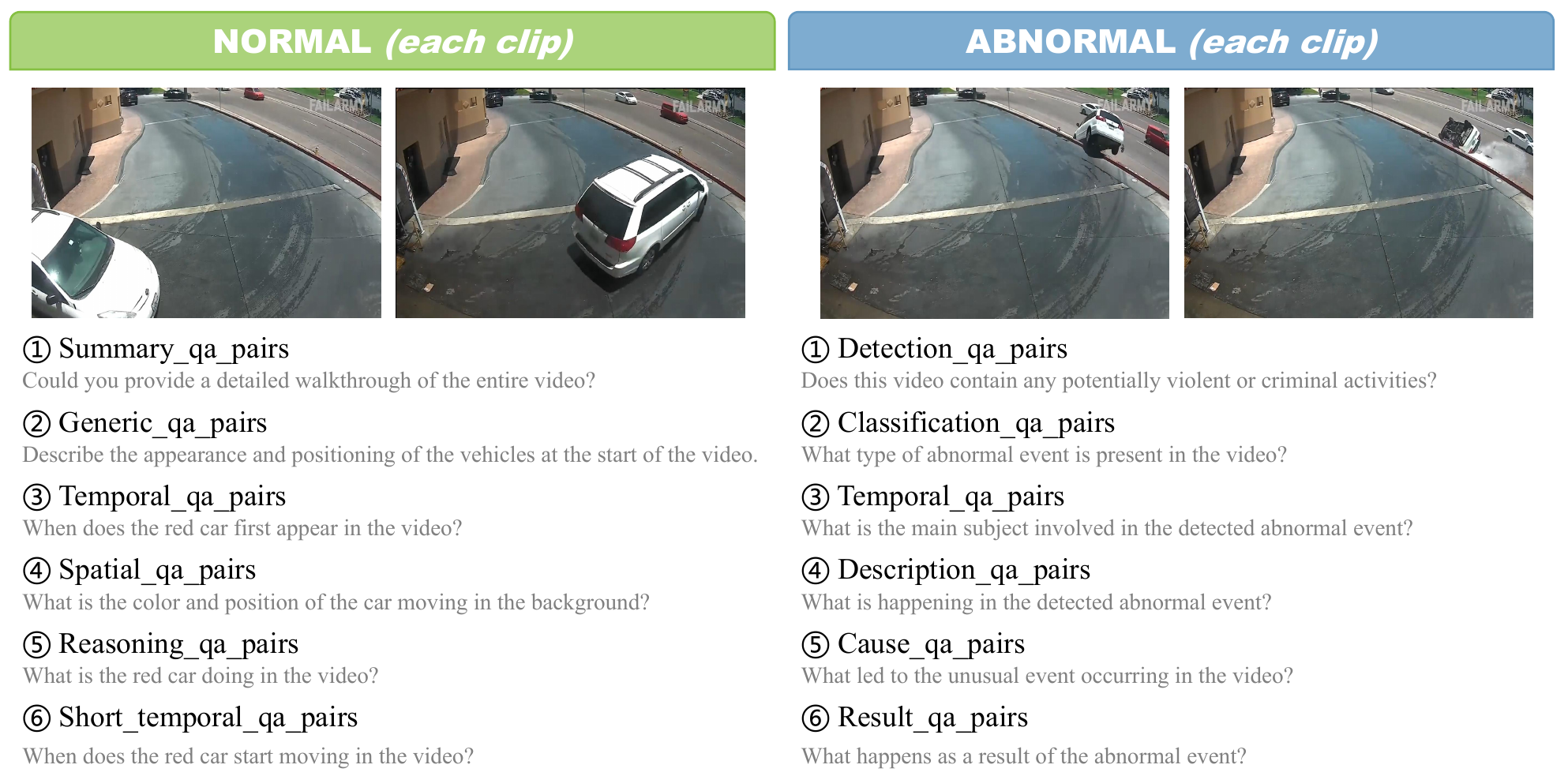}
       \caption{Normal and abnormal QA tasks.}
       \label{fig:sub-b}
    \end{subfigure}
    \vspace{-0.5em}
    \caption{Proposition of our SurveillanceVQA tasks.}
    \label{fig:main}
\end{figure}

\section{Introduction}

Surveillance videos have become pivotal data sources for smart cities~\cite{kashef2021smart}. In contrast to publicly available or social media videos, surveillance footage varies significantly in acquisition methods, content attributes, and application purposes~\cite{xu2016video,tsakanikas2018video}. They encompass diverse spatiotemporal conditions~\cite{Nawaratne2020SpatiotemporalAD,Sreenu2019IntelligentVS}, spanning day-night cycles, varied weather, and heterogeneous environments like streets~\cite{liang2023revealing,ma2019data}, shopping centers~\cite{arroyo2015expert}, and transportation hubs~\cite{ling2017case}, resulting in high data heterogeneity. Abnormal events captured are often sudden, low-frequency, and diverse, posing challenges for perception and modeling~\cite{liu2024generalized,Liu2023GeneralizedVA}.

Current computer vision tasks for surveillance videos primarily focus on the detection and classification of abnormal events, often relying on predefined event types and handcrafted features~\cite{pawar2019deep,zhou2019anomalynet,doshi2020any,al2024collaborative,zanella2024harnessing,wu2024open}. While such goal-oriented approaches can be effective in specific scenarios, they typically lack deeper semantic modeling of event progression, behavioral motivations, and environmental context~\cite{yuan2024surveillance}. This limits the potential of surveillance videos in intelligent urban governance, behavior prediction, and multimodal reasoning~\cite{pathirannahalage2024comprehensive}.

To address these limitations, our previous work, UCA~\cite{yuan2024towards,yuan2024surveillance}, has proposed a multimodal understanding framework for surveillance videos. It incorporates fine-grained language annotations and temporal markers, covering tasks such as moment localization, caption generation, and dense captioning. This framework extends the boundaries of traditional surveillance video analysis by enhancing its semantic expressiveness. However, UCA primarily focuses on descriptive tasks and lacks interactive question answering (QA) mechanisms, which makes it less aligned with recent trends in complex semantic understanding and reasoning within multimodal systems~\cite{kim2025visual}.

To further enhance the semantic reasoning capabilities of models in the surveillance domain, we introduce QA tasks to enable interactive and cognitively rich understanding as shown in Figure~\ref{fig:main}. Unlike descriptive tasks, QA tasks allow models to perform logical reasoning, causal inference, and complex semantic analysis on video content. To achieve this, we construct SurveillanceVQA-589K, a large-scale QA dataset specifically designed for surveillance videos. The dataset consists of four surveillance video datasets as video sources and contains approximately 589,000 question-answer pairs, including 12 QA types covering both normal and abnormal video content, such as factual summarization, behavior/spatial-temporal analysis, causal reasoning, anomaly detection, etc. This design aims to elevate video understanding to a higher cognitive level. 

We benchmark 8 Large Vision-Language Models (LVLMs) on SurveillanceVQA-589K, including variants like VideoLLaMa3~\cite{zhang2025videollama}, LLaVA series~\cite{zhang2024videoinstructiontuningsynthetic, zhang2024llavanextvideo, li2024llava}, Qwen2.5-VL series~\cite{Qwen2.5-VL}, and InternVL series~\cite{chen2024expanding}, ranging from lightweight 0.5B to general-purpose 7B models. Despite their success in open-domain tasks, current LVLMs demonstrate clear limitations in surveillance scenarios. Performance on complex tasks, such as causal inference and abnormal event analysis, remains poor, with most models scoring below the midpoint. Furthermore, while fine-tuning improves general understanding, it fails to significantly enhance structured reasoning tasks like anomaly detection and classification.

Our main contributions are as follows.
\begin{itemize}
    \item We construct SurveillanceVQA-589K, the largest surveillance video QA dataset to date, containing 589,380 QA pairs across 12 task types and 18 abnormal event categories, enabling comprehensive semantic understanding under both normal and abnormal conditions.
    \item We introduce a scalable hybrid annotation pipeline, combining human-aligned captions with LVLM-generated content, and propose a multi-dimensional evaluation framework tailored to surveillance video reasoning tasks, including contextual, temporal, and causal metrics.
    \item Our benchmark evaluates eight open-source LVLMs (0.5B–7B) and reveals systematic weaknesses in causal inference and anomaly understanding, offering a practical testbed for real-world applications such as intelligent security and video anomaly response systems.

\end{itemize}

\section{Related Works}

\subsection{Surveillance Video Analysis Benchmark }
In recent years, there has been a growing interest in the academic community toward understanding the content of surveillance videos, which has led to the development of several benchmark datasets to support research in this domain. Early datasets predominantly focused on anomaly detection in surveillance scenarios, including UCSD Ped1 and Ped2 \cite{li2013anomaly}, the Avenue dataset \cite{lu2013abnormal}, the Subway dataset \cite{adam2008robust}, the ShanghaiTech Campus dataset \cite{luo2017revisit}, UCF-Crime \cite{sultani2018real}, MEVA \cite{Corona_2021_WACV}, NWPU Campus dataset \cite{nwpu2023}, and MSAD \cite{msad2024}. While these datasets have laid a strong foundation for surveillance video analysis, the majority of these datasets still lack accompanying textual annotations, rendering them inadequate for comprehensive multimodal vision-language research. 
Notably, UCA \cite{yuan2024towards} distinguished itself from prior surveillance datasets through its rich linguistic annotations. However, the current version of UCA includes only approximately 20,000 manually labeled descriptions and lacks an interactive, question-answering (QA) based evaluation framework. Such a framework is increasingly recognized as a critical tool for assessing high-level reasoning, anomaly understanding, and semantic generalization in modern multimodal systems. 

To address this gap, we propose the construction of a novel QA-driven benchmark for surveillance video understanding. This benchmark is designed to enable interactive evaluation and foster deeper semantic reasoning over real-world surveillance video content. 

\subsection{Video-language Understanding Benchmark}

With the emergence of LVLMs~\cite{Li2024ASO,Cui2023ASO,Liu2023MMSafetyBenchAB}, traditional benchmarks have become increasingly inadequate in capturing the full range of model capabilities. Recent benchmarks aim to address this by incorporating diverse tasks~\cite{Wang2024ExploringTR,Li2025ASO}, multi-level granularity~\cite{Wang2024MultimodalNI}, and scalable evaluation protocols~\cite{Chen2024PCABenchEM}. Benchmarks such as OwlEval, MME~\cite{Fu2023MMEAC}, SEED-Bench~\cite{li2024seed}, MM-Vet~\cite{yu2023mm}, and MMBench~\cite{liu2024mmbench}, MVBench~\cite{li2024mvbench}, Vision-R1~\cite{Huang2025VisionR1IR}, and EMMA~\cite{Hao2025CanMR} cover a wide spectrum of tasks, from image captioning and reasoning to fact verification. These works propose multi-dimensional metrics—including linguistic consistency, semantic alignment, and visual grounding—to characterize model behavior more comprehensively. In the video domain, models must handle temporal dynamics and evolving semantics~\cite{Weng2024LongVLMEL,Chen2024ShareGPT4VideoIV}. Earlier benchmarks such as TVQA~\cite{Lei2018TVQALC} and Next-QA~\cite{Xiao2021NExTQANP} used multiple-choice formats to assess temporal localization and event comprehension. More recent work, such as VideoChatGPT~\cite{maaz2023video}, introduces multi-turn QA grounded in video input, emphasizing coherence and contextual consistency over extended interactions. FunQA~\cite{xie2024funqa} pushes reasoning further by evaluating a model’s ability to identify unexpected or humorous events, highlighting challenges in modeling incongruity and causal anomalies in temporal sequences. Svbench~\cite{Yang2025SVBenchAB} proposes temporal multi-turn QA chains, which are specifically for streaming video understanding. 


Our work extends these efforts by addressing the unique challenges of surveillance video analysis. Unlike open-domain or entertainment-based datasets, SurveillanceVQA-589K emphasizes real-world diverse abnormal events, spatiotemporal analysis, complex reasoning, etc.


\section{SurveillanceVQA-589K}

\begin{figure}
    \centering
    \includegraphics[width=1.\linewidth]{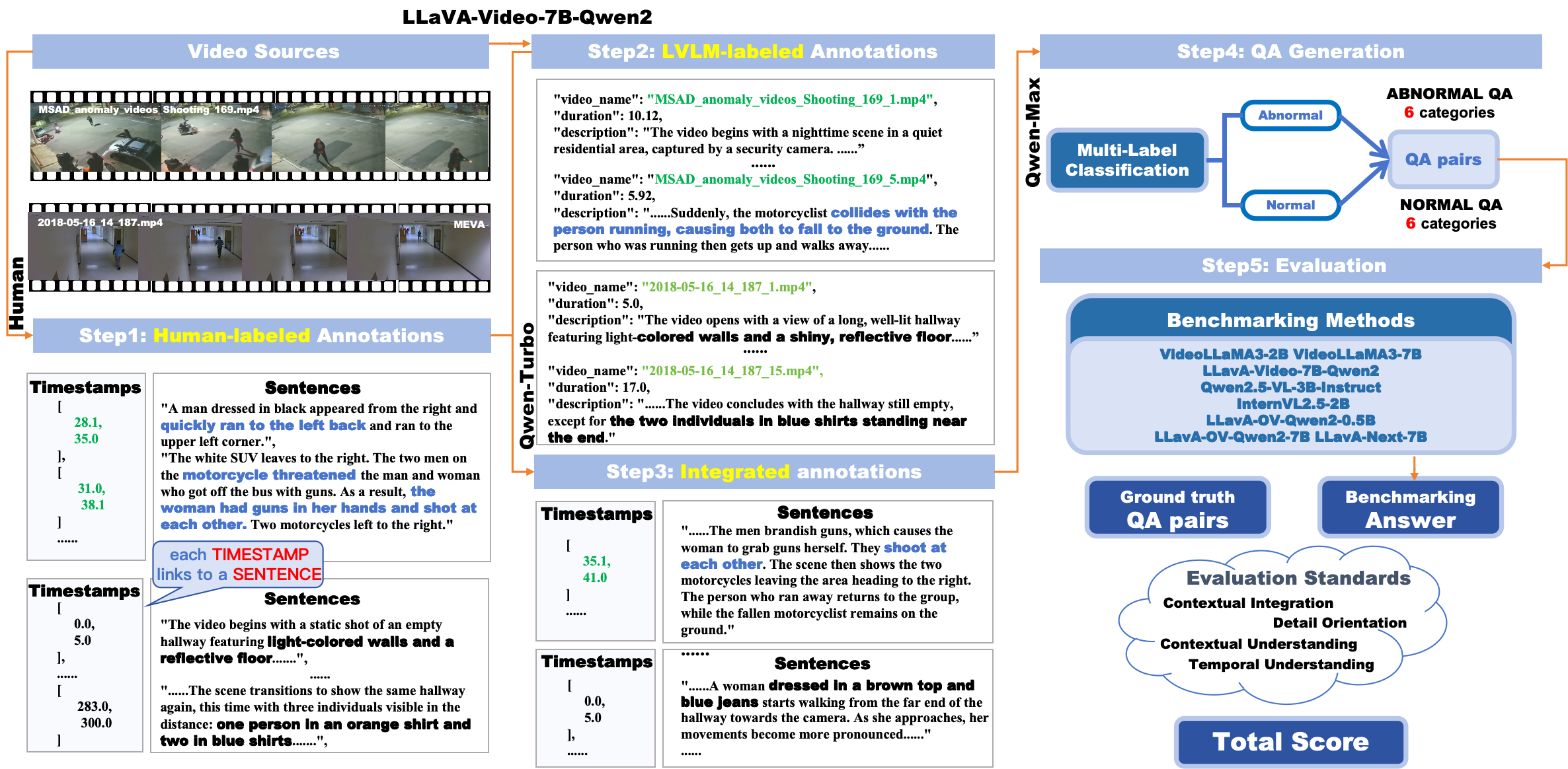}
    \caption{Our overall framework, including QA generation and evaluation.}
    \label{fig:enter-label}
\end{figure}

The SurveillanceVQA-589K dataset includes 31,548 video clips with textual annotations, 27,966 clips labeled as normal and 3,585 as anomalous, resulting in a total of 589,380 QA pairs. The following contents show the procedure of QA pairs generation, illustrated in Figure~\ref{fig:enter-label}. 

\subsection{Video Annotation Generation}
The examples of human-labeled, LVLM-labeled, integrated annotations are shown in Appendix~\ref{AppendixA}. 

\subsubsection{Human-labeled Annotation}

Following the annotation protocol established in UCA~\cite{yuan2024towards}, we extended manual annotation efforts to the MSAD, MEVA, and NWPU surveillance datasets. This process involved generating event-level captions that included both precise timestamps and detailed event descriptions.

To ensure annotation accuracy and consistency, we provided annotators with clear guidelines and comprehensive training, following the protocol established in UCA~\cite{yuan2024towards}. The annotation was carried out by a team of technically proficient annotators who were fairly compensated according to local wage standards. The process was supervised by AI researchers who regularly reviewed and validated the outputs to ensure both quality and ethical compliance. The entire annotation phase spanned approximately one month and resulted in a high-quality corpus. We then integrated this newly annotated data with existing annotations from UCA, completing the manual annotation collection with a total of 31,548 sentence-level annotations accompanied by precise timestamps.

\subsubsection{LVLM-labeled Annotation}

Subsequently, using the timestamp information obtained during the manual annotation phase, we employed the video processing toolkit MoviePy to automatically segment the original videos and extract the corresponding short clips. We then utilized the powerful multimodal model LLaVA-Video-7B-Qwen2~\cite{zhang2024videoinstructiontuningsynthetic} to perform in-depth analysis on each segmented clip, generating detailed descriptions. As a result, we obtained 31,548 segment-level annotations produced by the LVLM.

\subsubsection{Integrated Human-LVLM Annotations}


\begin{wrapfigure}[9]{r}{0.5\textwidth}
\vspace{-1em}
\includegraphics[width=0.5\columnwidth]{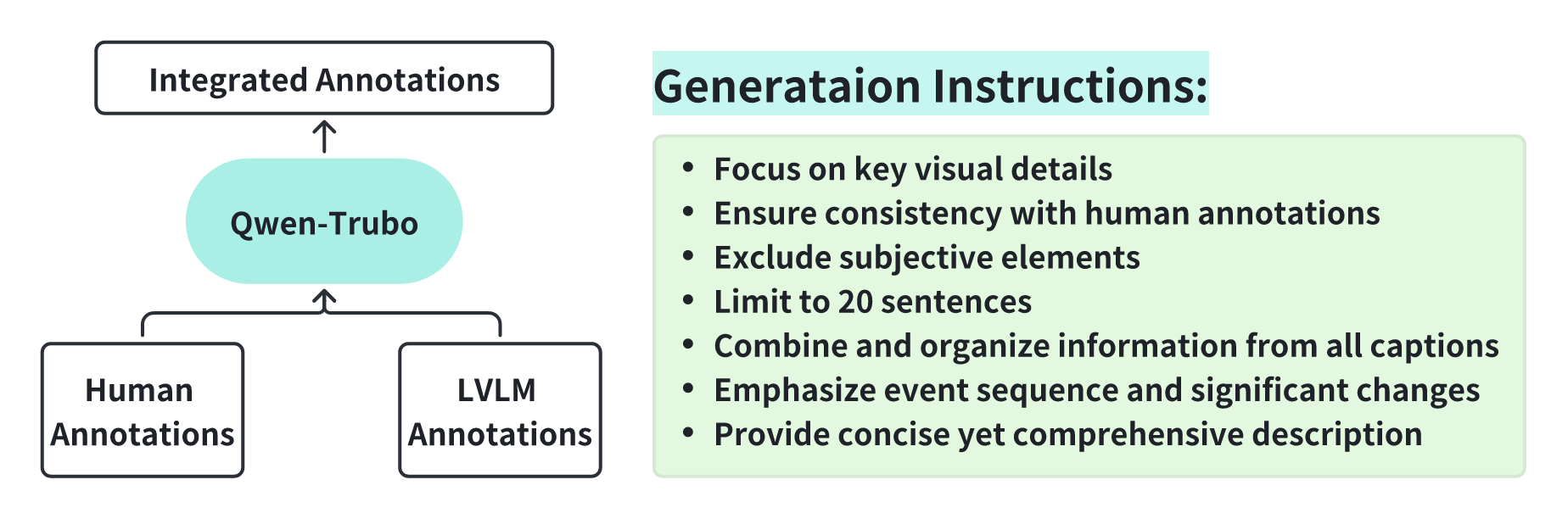}
\caption{Procedure of integrating annotations.}
\label{fig:prompts}
\end{wrapfigure}

In this stage, we performed a deep integration of the high-quality information obtained from manual annotations with the rich segment-level descriptions generated by the large multimodal model LLaVA-Video. The objective of this integration was to combine the precision of human annotations with the diversity and depth of model-generated descriptions. The guiding principle for prompt design was to preserve the semantic integrity of human annotations while enriching them with complementary information from the model output. To facilitate this process, we introduced Qwen-Turbo~\cite{qwen2.5}, an advanced language model, to leverage its natural language processing capabilities for comprehensive analysis, redundancy elimination, and content optimization. Specifically, Qwen-Turbo was tasked with identifying and resolving redundant or inconsistent expressions, while enhancing semantic richness and logical coherence. This resulted in more fluent, structured, and contextually aligned event-level descriptions for each video clip. 

Through this integration, we obtained 31,548 refined textual annotations. The procedure for integrating human and LVLM annotations, along with the generation prompt, is illustrated in Figure~\ref{fig:prompts}.


\subsection{Automatic QA Generation}

After obtaining the clip-level descriptions, we proceeded to generate QA pairs.




\subsubsection{QA Category Definition}

Specifically, we employed Qwen-Max~\cite{qwen2.5} to analyze the annotation data and classify each video segment as either normal or abnormal. 

After obtaining two distinct sets of video clips (normal vs. abnormal) along with their corresponding textual annotations, we designed category-specific prompts to guide the generation of QA pairs. For normal clips, the prompts were crafted to elicit a comprehensive understanding of video content, focusing on global scene descriptions, temporal sequencing, spatial detail extraction, and behavioral inference. These prompts support general video comprehension and open-ended QA tasks. In contrast, for abnormal clips, the prompts emphasized event detection, anomaly type classification, subject identification, detailed incident descriptions, and causal reasoning. This design is tailored to the specific requirements of surveillance anomaly detection and incident-level semantic analysis.

This balanced approach ensures both efficiency and semantic richness across diverse surveillance contexts. The detailed prompts are summarized in Table~\ref{Designed_prompts} of the Appendix~\ref{AppendixB}.


\subsubsection{QA Generation}
Based on the customized prompts, we employed Qwen-Max~\cite{qwen2.5} to generate QA pairs for each video clip, following different generation strategies for normal and abnormal categories. For normal QA tasks, six QA types~(Summary, Generic, Temporal,  Short Temporal, Spatial, and Reasoning QA) were defined, with three QA pairs generated per type. For abnormal QA tasks, the other six QA types (Detection, Classification, Temporal, Description, Cause, and Result QA) were applied, but with one QA pair generated per type to emphasize critical semantic cues. Notably, all of the video clips (including normal and abnormal clips) are suitable for the normal QA tasks with their corresponding normal QA pairs, while only abnormal clips are suitable for the abnormal QA tasks with particularly defined abnormal QA pairs.

The defined QA categories are illustrated in Figure~\ref{fig:sub-b}, and detailed descriptions of each QA task are provided in the Appendix~\ref{AppendixB}.



\subsection{Data Statistics}
\subsubsection{Overall statistics}

\begin{wraptable}[9]{r}{0.62\textwidth}
\vspace{-1em}
\centering
\caption{Overall data statistics of SurveillanceVQA-589K.}
\resizebox{0.62\textwidth}{!}{
\begin{tabular}{lcccccr}
\toprule
\multirow{2}{*}{Dataset} & \multirow{2}{*}{\begin{tabular}[c]{@{}c@{}}Number of\\ Videos\end{tabular}} & \multirow{2}{*}{\begin{tabular}[c]{@{}c@{}}Total Video\\ Duration\end{tabular}} & \multirow{2}{*}{\begin{tabular}[c]{@{}c@{}}Number of Text\\ Annotations\end{tabular}} & \multicolumn{2}{c}{Number of Segments} & \multirow{2}{*}{\begin{tabular}[c]{@{}c@{}}QA\\ Pairs\end{tabular}} \\ 
\cmidrule(lr){5-6}
 &  &  &  & Normal & Abnormal &  \\ 
\midrule
MSAD & 201 & 4.23h & 1783 & 1417 & 366 & 34290 \\ 
MEVA & 720 & 16.76h & 2057 & 2044 & 13 & 37104 \\ 
NWPU & 255 & 16.29h & 4166 & 4121 & 45 & 75258 \\ 
UCA & 1854 & 121.9h & 23542 & 20380 & 3162 & 442728 \\ 
\midrule
\multicolumn{1}{l}{\textbf{Total}} & \textbf{3030} & \textbf{159.18h} & \textbf{31548} & \textbf{27962} & \textbf{3586} & \textbf{589380} \\
\bottomrule
\end{tabular}}
\label{table-four-numbers}
\end{wraptable}

Table~\ref{table-four-numbers} presents statistical information for the SurveillanceVQA-589K (containing MSAD, MEVA, NWPU, and UCA), including the number of videos, total duration, textual annotations, segmented clips (categorized as normal and abnormal), and the total number of QA pairs. More details are presented in Appendix~\ref{AppendixC}.


\subsubsection{Categorization Statistics}

In terms of dataset partitioning, we split the dataset at the clip level rather than by entire videos, using an 8/2 ratio for the training and testing sets. Due to the difficulty of annotating every second of footage, the total duration of the training and testing sets is slightly shorter than that of the original raw video collection. Table~\ref{distribution of normal QA pairs} presents the distribution of QA pairs for normal and abnormal events across the training and testing sets. This table reports the number of QA pairs generated for both event types and illustrates how their distribution varies between subsets for each dataset. Here, the normal category refers to non-anomalous clips, while the abnormal category corresponds to anomalous clips, which span 18 distinct abnormal event classes as detailed in the Appendix~\ref{AppendixC}.


\begin{table}
\centering
\caption{Distribution of QA pairs for normal and abnormal events across the training and testing sets}
\resizebox{0.8\textwidth}{!}{
\begin{tabular}{l *{8}{c}}
\hline
& \multicolumn{2}{c}{\textbf{MEVA}} & \multicolumn{2}{c}{\textbf{MSAD}} & \multicolumn{2}{c}{\textbf{NWPU}} & \multicolumn{2}{c}{\textbf{UCA}} \\
\hline
& \textbf{Abnormal} & \textbf{Normal} & \textbf{Abnormal} & \textbf{Normal} & \textbf{Abnormal} & \textbf{Normal} & \textbf{Abnormal} & \textbf{Normal} \\
\hline
Test & 72 & 7362 & 1776 & 5112 & 240 & 14850 & 15192 & 73368 \\
Train & 240 & 29430 & 7008 & 20394 & 840 & 59328 & 60696 & 293472 \\
\hline
\end{tabular}
}
\label{distribution of normal QA pairs}
\end{table}

\subsubsection{Comparisons with existing datasets}


These comparisons, as shown in Table~\ref{Comparison with existing datasets}, reveal a consistent trend: early video QA datasets are limited in scale, typically containing 200 to 900 videos and 2,000 to 4,000 QA pairs, and primarily focus on normal events with fixed question formats. Moreover, most existing datasets rely exclusively on either human annotations or LLM-generated content, with few employing a hybrid approach that leverages both. UCVL~\cite{Chen2025ABF} serves as an intermediate example by exploring abnormal event QA, yet it still falls short in terms of scale, diversity, and task coverage compared to our SurveillanceVQA-589K dataset.

\begin{table}[!ht]
    \centering
    \caption{Overall comparisons with existing datasets.}
    \resizebox{0.8\textwidth}{!}{
    \begin{tabular}{lcccccc}
        \toprule
        \textbf{Aspect} & \textbf{MVBench~\cite{li2024mvbench}} & \textbf{Video-MME~\cite{Fu2024VideoMMETF}} & \textbf{MMB-Video~\cite{Fang2024MMBenchVideoAL}} & \textbf{UCVL~\cite{Chen2025ABF}} & \textbf{SurveillanceVQA-589K} \\
        \midrule
        \textbf{Videos} & 200 & 900 & 600 & 1699 & 3030 \\
        \textbf{QA Pairs} & 4000 & 2700 & 2000 & 16990 & 589380 \\
        \textbf{Content} & Normal & Normal & Normal & Anomaly & Normal\&Anomaly \\
        \textbf{QA Forms} & MCQ & MCQ & Open-ended & MCQ\&Open-ended &Open-ended \\
        \textbf{Generation} & LLM & Human & Human & LLM & LLM\&Human \\
        \textbf{Evaluation} & Matching & Matching & LLM & Matching\&LLM & LLM \\
        \bottomrule
    \end{tabular}
    }
    \label{Comparison with existing datasets}
\end{table}

\section{Experiments on SurveillanceVQA-589K}
We provide the evaluations, baselines, settings, and experimental results in this section. More detailed experimental results and explanations have been given in the Appendix~\ref{AppendixD}.

\subsection{Evaluation Design}
We follow VideoGPT+~\cite{maaz2024videogpt+} to design our evaluation criteria. During the evaluation phase, we adopt LLM-based evaluation strategy, using an open API with GLM-4-Flash~\cite{glm2024chatglm}. This approach ensures that the evaluation process not only considers the semantic consistency of the answers but also leverages the interpretive and reasoning capabilities of advanced language models. 

This evaluation framework provides a comprehensive assessment of model-generated answers across four key dimensions. Contextual Integration (CI) measures whether the answer accurately reflects the factual content of the video, avoiding errors or misinterpretations. Detail Orientation (DO) assesses the inclusion of specific and complete key elements. Contextual Understanding (CU) evaluates the alignment of the answer with the overall narrative and emotional tone of the video. Temporal Understanding (TU) focuses on the correctness of event sequences and time-related logic. Each dimension is rated on a 0–5 integer scale, with 5 indicating full accuracy and relevance, and 0 indicating a completely incorrect response. To calculate Average Score~(Avg), individual scores are normalized by multiplying each by 0.25 and summing the results. This scoring scheme enables fine-grained, quantitative evaluation of model performance across multiple facets of video understanding.

\subsection{Baselines and Settings}

We evaluate 8 open-source video-language models, including the VideoLLaMA3~\cite{zhang2025videollama}, InternVL2.5~\cite{chen2024expanding}, LLavA-OV-Qwen2~\cite{li2024llava}), LLaVA-Video-Qwen2~\cite{zhang2024videoinstructiontuningsynthetic}, and Qwen2.5-VL-Instruct series~\cite{Qwen2.5-VL}, with parameter sizes ranging from 0.5B to 7B. Each model performs inference on one question at a time to prevent information leakage between questions. We select Qwen2.5-VL-Instruct-3B for LoRA fine-tuning, which offers a good balance between parameter size and performance, and conduct the fine-tuning for one epoch on our training set. Evaluation is conducted on our test set, and all experiments are carried out on an NVIDIA RTX 4090 GPU. Table~\ref{tab:model_overview} shows the characteristics of these evaluated open-source LVLMs.

\begin{table}[ht]
\centering
\caption{Overview of Evaluated Open-Source Video Models}
\resizebox{.85\textwidth}{!}{
\begin{tabular}{ll}
\toprule
\textbf{Model Name}   & \textbf{Key Features} \\
\midrule
Video-LLaMA3-2B/7B~\cite{zhang2025videollama}& Uses any-resolution vision tokenization and differential frame pruner to reduce \\ & information loss and computation cost. Trained on high-quality video data. \\ \midrule

InternVL2.5-2B~\cite{chen2024expanding} & Based on the ViT-MLP-LLM framework. Incorporates dynamic high-resolution \\ & representations, Progressive scaling Strategy, and Chain-of-Thought (CoT) reasoning.  \\ \midrule

LLavA-OV-Qwen2-0.5B/7B~\cite{li2024llava}) & Multimodal model capable of understanding single images, multiple images, \\ & and videos. Supports cross-modal transfer learning. \\
\midrule
LLaVA-Video-7B-Qwen2~\cite{zhang2024videoinstructiontuningsynthetic} & Trained only on text-image data with AnyRes technology. Fine-tuned \\ &  on LLaVA-Video-178K for enhanced video instruction understanding.  \\
\midrule
Qwen2.5-VL-3B-Instruct~\cite{Qwen2.5-VL} & Strong instruction-following capabilities across text, image, \\ & and video. Improved cross-modal alignment and QA generation. \\
\bottomrule
\end{tabular}}
\label{tab:model_overview}
\end{table}

\subsection{Results on SurveillanceVQA-589K}

\subsubsection{Analysis of overall LVLMs performance}

Table~\ref{tab:model_performance} presents the evaluation results of various LVLMs across five key dimensions: CI, DO, CU, TU, along with Avg. LLaVA-Video-7B-Qwen2 achieved the highest overall performance, with an average score of 2.72. This strong performance can be attributed to its integration of the LLaVA and Qwen2 architectures and its use of the AnyRes technique, which enhances image-to-video reasoning. This combination enables the model to maintain high cross-modal consistency and accuracy, especially in capturing the temporal flow and nuanced context of video events. In contrast, InternVL2.5-2B demonstrated the lowest overall average, with notably weak scores in CI and DO. 
\begin{wraptable}[17]{l}{0.5\textwidth}
\centering
\caption{Model performance averaged on different QA tasks across five evaluation dimensions. $^{\dagger}$ represents our finetuned LVLMs.}
\resizebox{0.5\textwidth}{!}{
\begin{tabular}{lccccc}
\toprule
\textbf{Method} & \textbf{CI} & \textbf{DO} & \textbf{CU} & \textbf{TU} & \textbf{Avg} \\
\midrule
LLavA-OV-Qwen2-0.5B & 2.89 & 2.62 & 2.89 & 2.64 & 2.76  \\
\midrule
InternVL2.5-2B & 1.77 & 1.72 & 1.97 & 1.72 & 1.79 \\
\midrule
VideoLLaMA3-2B & 2.82 & 2.55 & 2.83 & 2.58 & 2.69 \\
\midrule
\rowcolor{gray!30}Qwen2.5-VL-3B-Instruct & 2.70 & 2.54 & 2.72 & 2.45 & 2.60 \\
\midrule
LLaVA-NeXT-Video-7B & 2.78 & 2.62 & 2.80 & 2.50 & 2.68 \\
\midrule
LLavA-OV-Qwen2-7B & 3.15 & 2.85 & 3.12 & 2.89 & 3.00 \\
\midrule
LLaVA-Video-7B-Qwen2 & 3.17 & 2.85 & 3.15 & 2.92 & 3.02 \\
\midrule
VideoLLaMA3-7B & 2.93 & 2.67 & 2.93 & 2.70 & 2.80 \\
\midrule
\rowcolor{gray!30}$\text{Qwen2.5-VL-3B-Instruct}^{\dagger}$ & 2.83 & 2.71 & 2.83 & 2.58 & 2.74\\
\bottomrule
\end{tabular}
}
\label{tab:model_performance}
\end{wraptable}
While its surface-level integration of context and detail may lag, it retains potential in complex reasoning scenarios due to a Chain-of-Thought (CoT) reasoning mechanism. Interestingly, the LLavA-OV-Qwen2-0.5B model, with just 0.5B parameters, achieved a competitive average score of 2.76. This demonstrates that smaller-scale models can still deliver strong video comprehension, thanks to efficient modality transfer learning between visual and textual inputs. Its performance affirms that parameter size is not the sole determinant of effectiveness in multimodal video tasks. The Qwen2.5-VL-3B-Instruct model also showed robust performance. Designed for multimodal instruction-following, this model emphasizes cross-modal alignment and optimized question-answer generation, equipping it to handle a variety of video, image, and text tasks with precision. Its balanced size and architecture strike an effective trade-off between computational efficiency and capability in understanding context. Finally, Video-LLaMA3-2B/7B incorporates advanced features like Any-resolution Vision Tokenization and the Differential Frame Pruner, which improve video representation quality and processing efficiency across varied resolutions and temporal segments.


\subsubsection{Analysis of LVLMs performance across normal QA tasks}
Table~\ref{tab:combined_qa_results} presents the performance of several LVLMs on a range of QA tasks, including Summary, Generic, Temporal, Short Temporal, Spatial, and Reasoning QA. The evaluation is conducted on both normal video clips (blue) and abnormal video clips (green), with performance metrics reported for each task. Among the models, LLaVA-OV-Qwen2-7B demonstrates the highest overall performance, indicating robust capabilities in handling spatial and reasoning-based questions. In contrast, InternVL2.5-2B exhibits the lowest performance across most tasks, with a Summary QA score of 0.56/0.37, suggesting limitations in processing video-based QA tasks effectively. 

A notable trend is the higher performance on normal video clips compared to abnormal ones across all models and tasks. For instance, VideoLLLaMA3-7B scores 3.01/2.70 in Generic QA for normal versus abnormal clips, respectively, highlighting the challenge posed by abnormal video contexts. 

\subsubsection{Analysis of LVLMs performance across abnormal QA tasks}

Table~\ref{tab:combined_qa_results} presents the performance of various vision-language models on abnormal QA tasks, including Detection QA, Classification QA, Subject QA, Description QA, Cause QA, and Result QA, evaluated exclusively on abnormal video clips with all values reported in Brown. In QA tasks involving abnormal video scenarios, LVLMs demonstrate notable performance variations and task-specific capabilities. Overall, the LLaVA-OV-Qwen2 series models excel in Detection tasks, with the 0.5B variant achieving top scores on CI and CU metrics. For Classification tasks, both LLaVA-Video-7B and LLaVA-OV-Qwen2-7B showcase superior performance across different evaluation criteria. The LLaVA-Video-7B model consistently demonstrates advantages in Subject and Description type questions. Notably, all models perform substantially weaker on higher-order reasoning tasks such as Cause and Result inference, with scores generally falling below the midpoint threshold. LLaVA-NeXT-7B performs relatively better in causal reasoning while LLaVA-Video-7B shows slight advantages in result inference. These patterns indicate that current vision-language models still face significant challenges in understanding causal relationships and reasoning about complex video content, particularly in abnormal video scenarios. Additionally, model scale appears to influence performance, with larger 7B-parameter models generally outperforming smaller variants across most abnormal video QA tasks, though distinct specializations emerge across different task categories.


Overall, the abnormal QA scores in Table~\ref{tab:combined_qa_results} are much lower than normal QA scores in Table~\ref{tab:combined_qa_results}, indicating more difficulties in abnormal QA interacting and reasoning.

\begin{table*}[htbp]
\centering
\caption{Performance of different vision-language models across QA tasks. {\color{normalblue}Blue}: normal QA tasks on normal video clips, {\color{abnormalgreen}Green}: normal QA tasks on abnormal video clips, {\color{browncolor}Brown}: abnormal QA tasks on abnormal video clips. $^{\dagger}$ represents our finetuned LVLMs.}
\Huge{
\resizebox{0.95\textwidth}{!}{
\renewcommand\arraystretch{1.8}
\begin{tabular}{c|cccccc|cccccc}
\toprule
\textbf{Model} & \textbf{Summary} & \textbf{Generic} & \textbf{Temporal} & \textbf{Short Temporal} & \textbf{Spatial} & \textbf{Reasoning} & \textbf{Detection} & \textbf{Classification} & \textbf{Subject} & \textbf{Description} & \textbf{Cause} & \textbf{Result} \\
\midrule
LLavA-OV-Qwen2-0.5B & \color{normalblue}2.76/\color{abnormalgreen}2.43 & \color{normalblue}2.78/\color{abnormalgreen}2.52 & \color{normalblue}2.62/\color{abnormalgreen}2.43 & \color{normalblue}2.60/\color{abnormalgreen}2.45 & \color{normalblue}3.10/\color{abnormalgreen}3.12 & \color{normalblue}2.93/\color{abnormalgreen}2.69 & \color{browncolor}2.94 & \color{browncolor}2.47 & \color{browncolor}2.69 & \color{browncolor}2.43 & \color{browncolor}1.67 & \color{browncolor}1.60 \\
\midrule
InternVL2.5-2B&
\color{normalblue}0.56/\color{abnormalgreen}0.37 & \color{normalblue}2.20/\color{abnormalgreen}1.93 & \color{normalblue}1.89/\color{abnormalgreen}1.68 & \color{normalblue}2.08/\color{abnormalgreen}1.95 & \color{normalblue}1.92/\color{abnormalgreen}1.89 & \color{normalblue}2.41/\color{abnormalgreen}2.26 & \color{browncolor}1.88 & \color{browncolor}1.12 & \color{browncolor}1.46 & \color{browncolor}0.55 & \color{browncolor}0.64 & \color{browncolor}0.74 \\
\midrule
VideoLLaMA3-2B&
\color{normalblue}2.49/\color{abnormalgreen}2.00 & \color{normalblue}2.84/\color{abnormalgreen}2.55 & \color{normalblue}2.73/\color{abnormalgreen}2.49 & \color{normalblue}2.61/\color{abnormalgreen}2.44 & \color{normalblue}2.97/\color{abnormalgreen}2.99 & \color{normalblue}2.89/\color{abnormalgreen}2.66 & \color{browncolor}1.87 & \color{browncolor}2.03 & \color{browncolor}2.46 & \color{browncolor}1.89 & \color{browncolor}1.37 & \color{browncolor}1.18 \\
\midrule
\rowcolor{gray!30}Qwen2.5-VL-3B-Instruct &
\color{normalblue}2.20/\color{abnormalgreen}1.49 & \color{normalblue}2.66/\color{abnormalgreen}2.17 & \color{normalblue}2.66/\color{abnormalgreen}2.26 & \color{normalblue}2.69/\color{abnormalgreen}2.31 & \color{normalblue}2.86/\color{abnormalgreen}2.75 & \color{normalblue}3.01/\color{abnormalgreen}2.70 & \color{browncolor}1.85 & \color{browncolor}2.19 & \color{browncolor}2.24 & \color{browncolor}1.74 & \color{browncolor}1.32 & \color{browncolor}1.13 \\
\midrule
LLaVA-NeXT-7B &
\color{normalblue}2.17/\color{abnormalgreen}1.72 & \color{normalblue}2.93/\color{abnormalgreen}2.60 & \color{normalblue}2.56/\color{abnormalgreen}2.28 & \color{normalblue}2.66/\color{abnormalgreen}2.48 & \color{normalblue}2.95/\color{abnormalgreen}2.95 & \color{normalblue}3.06/\color{abnormalgreen}2.81 & \color{browncolor}2.32 & \color{browncolor}2.56 & \color{browncolor}2.59 & \color{browncolor}2.11 & \color{browncolor}1.96 & \color{browncolor}1.65 \\
\midrule
LLavA-OV-Qwen2-7B&
\color{normalblue}3.10/\color{abnormalgreen}2.79 & \color{normalblue}3.08/\color{abnormalgreen}2.86 & \color{normalblue}2.95/\color{abnormalgreen}2.81 & \color{normalblue}2.79/\color{abnormalgreen}2.66 & \color{normalblue}3.31/\color{abnormalgreen}3.34 & \color{normalblue}3.08/\color{abnormalgreen}2.89 & \color{browncolor}2.53 & \color{browncolor}2.60 & \color{browncolor}2.68 & \color{browncolor}2.54 & \color{browncolor}1.46 & \color{browncolor}1.55 \\
\midrule
LLaVA-Video-7B & 
\color{normalblue}3.05/\color{abnormalgreen}2.85 & \color{normalblue}2.99/\color{abnormalgreen}2.80 & \color{normalblue}2.85/\color{abnormalgreen}2.77 & \color{normalblue}2.78/\color{abnormalgreen}2.71 & \color{normalblue}3.47/\color{abnormalgreen}3.51 & \color{normalblue}3.25/\color{abnormalgreen}3.05 & \color{browncolor}1.92 & \color{browncolor}2.59 & \color{browncolor}2.76 & \color{browncolor}2.76 & \color{browncolor}1.52 & \color{browncolor}1.81 \\
\midrule
VideoLLaMA3-7B&
\color{normalblue}2.68/\color{abnormalgreen}2.29 & \color{normalblue}3.01/\color{abnormalgreen}2.70 & \color{normalblue}2.81/\color{abnormalgreen}2.55 & \color{normalblue}2.65/\color{abnormalgreen}2.45 & \color{normalblue}3.06/\color{abnormalgreen}3.06 & \color{normalblue}2.99/\color{abnormalgreen}2.78 & \color{browncolor}2.03 & \color{browncolor}1.53 & \color{browncolor}2.66 & \color{browncolor}2.01 & \color{browncolor}1.75 & \color{browncolor}1.27 \\
\midrule
\rowcolor{gray!30}$\text{Qwen2.5-VL-3B-Instruct}^{\dagger}$ & \color{normalblue}2.62/\color{abnormalgreen}1.99 & \color{normalblue}2.76/\color{abnormalgreen}2.36 & \color{normalblue}2.84/\color{abnormalgreen}2.48 & \color{normalblue}2.93/\color{abnormalgreen}2.59 & \color{normalblue}2.89/\color{abnormalgreen}2.89 & \color{normalblue}3.06/\color{abnormalgreen}2.85 & \color{browncolor}1.58 & \color{browncolor}2.13 & \color{browncolor}2.56 & \color{browncolor}2.06 & \color{browncolor}1.77 & \color{browncolor}1.22 \\
\bottomrule
\end{tabular}
}
}
\label{tab:combined_qa_results}
\end{table*}

\subsection{Comprehensive analysis}


\subsubsection{Analysis across different LVLMs and used techniques.} 

Smaller models, such as LLavA-OV-Qwen2-0.5B, excel in CU and DO but struggle with more complex tasks like TU. These models perform well in simpler scenarios but face limitations in long-duration tracking and anomaly detection. Medium models, like Qwen2.5-VL-3B-Instruct, provide a good balance between performance and efficiency, particularly in handling temporal sequences and contextual integration. Larger models, such as LLaVA-Video-7B-Qwen2 and Video-LLaMA3-7B, excel in temporal reasoning and anomaly detection, benefiting from advanced architectures and video-specific fine-tuning.

The experimental results indicate that fine-tuning yields modest performance gains. Fine-tuning models on specific monitoring video data significantly improves their performance in particular environments, helping them better understand context and identify anomalous events. However, fundamentally improving the model’s capability to align visual and textual information remains the key to achieving substantial progress. The strengths of these models can largely be attributed to their design choices. Techniques like Differential Frame Pruning and Progressive Scaling help the models handle long video sequences more efficiently while retaining important temporal details. Techniques like Any-Resolution Vision Tokenization, are capable of processing low-resolution video efficiently, suitable for monitoring applications where video quality may vary. 

\subsubsection{Analysis across QA types and video contexts.}

The performance of vision-language models varies significantly across different QA types and video contexts. Normal video QA tasks, such as Summary, Generic, Spatial, and Reasoning QA, generally achieve higher scores across all models. For instance, models like LLaVA-OV-Qwen2-7B and LLaVA-Video-7B consistently score above 3.0 on Spatial and Generic QA, indicating strong capabilities in visual description and object localization in well-structured scenes. In contrast, abnormal video QA tasks, especially Cause and Result QA, remain highly challenging. Across all models, scores for Cause QA fall below 2.0, with InternVL2.5-2B as low as 0.64, and the best-performing model (LLaVA-NeXT-7B) only reaching 1.96. This suggests a widespread limitation in causal reasoning, especially under chaotic or low-frequency events like violence or accidents.

Moreover, models generally perform better on normal video clips than on abnormal ones, even for the same QA types. This indicates that scene stability significantly influences model comprehension.



\textbf{Analysis of finetuned models.}

After LoRA fine-tuning, $\text{Qwen2.5-VL-3B-Instruct}^{\dagger}$ achieved an average score of 2.74, compared to 2.60 from the original model, a relative improvement of approximately 5.4\%. Notable gains were observed in CI (from 2.70 to 2.83) and DO (from 2.54 to 2.71), indicating enhanced ability in understanding scene context and capturing key details. While fine-tuning improved the model's performance on general understanding tasks, it has limited impact on anomaly-specific QA, particularly for Detection and Classification questions. This highlights a limitation of current fine-tuning approaches: they do not adequately enhance a model’s ability to produce structured, domain-specific expressions required for precise anomaly identification and classification.

\section{Ethical Considerations}

In this study, we strictly adhere to the ethical guidelines published by NeurIPS, addressing key ethical considerations throughout our data construction, model evaluation, and sharing processes. The video datasets used (such as MEVA, MSAD, NWPU, and UCF) are all publicly available and intended for research purposes, with no personally identifiable information involved, ensuring the legality and compliance of data sources. To mitigate potential biases, we incorporated diverse video scenarios and carefully crafted prompts during the QA generation and evaluation stages, aiming to reduce cultural or contextual bias inherent in language models or the original datasets. Moreover, we explicitly oppose any unauthorized use of our dataset or methods (e.g., for abusive surveillance or discriminatory profiling), and we deliberately avoided including high-risk content such as facial recognition. During data annotation, we employed a collaborative pipeline combining human annotators and large language models (e.g., Qwen-Max), with humans responsible for event segmentation and verification, and fair compensation provided in accordance with local wage standards. We plan to release our dataset and code under a research-friendly license with clear ethical usage guidelines, promoting responsible practices in multimodal research.

\section{Conclusion}

In this study, we introduce SurveillanceVQA-589K, the largest open-ended video QA benchmark tailored specifically to real-world surveillance scenarios. The dataset contains 589,380 QA pairs spanning 12 cognitively diverse task types across both normal and abnormal surveillance video contexts. We propose a hybrid annotation pipeline that combines human-aligned captions with LVLM-assisted QA generation, enabling high-quality, scalable annotation. We benchmark eight state-of-the-art open-source LVLMs ranging from 0.5B to 7B parameters. Our experiments reveal that while these models demonstrate promising performance on general understanding tasks (e.g., Summary, Spatial, and Qeneric QA on normal videos), they struggle significantly with complex semantic reasoning, particularly in anomaly-specific tasks such as causal inference and result prediction indicating a clear performance bottleneck in high-level temporal and logical reasoning. We also examine the impact of fine-tuning through LoRA on a 3B model. While fine-tuning yields moderate gains on general tasks, it provides limited improvement in structured anomaly detection and classification, highlighting that current parameter-efficient tuning approaches are insufficient for enabling domain-specific, structured expression. Overall, this work provides a comprehensive testbed for evaluating multimodal models in realistic surveillance settings. Our findings underscore the pressing need for LVLMs to develop stronger causal reasoning, temporal modeling, and structured response generation capabilities. 



\bibliographystyle{plainnat}
\bibliography{ref}


\newpage
\appendix

\section{Annotation Data Examples}~\label{AppendixA}

\begin{lstlisting}
Example of manually annotations:

"2018-03-15_10_131": {
    "timestamps": [
        [3.2, 7.7],
        [50.0, 54.0],
        [114.0, 118.0],
        [148.0, 156.0],
        [188.0, 194.0]
    ],
    "sentences": [
        "A man in a hat goes down to the underground passage.",
        "A cuckold man came up from the tunnel crossing.",
        "A woman with long hair carried a bag down the underground passage.",
        "A blue dress white hair man net underground passage down",
        "Two men came up from the underpass"
    ]
}
\end{lstlisting}

\begin{lstlisting}
Example1 of Detailed Description by LLavA-Next:
{
    "video_name": "2018-03-15_10_131_5.mp4",
    "duration": 6.0,
    "description": "The video opens with a view of a concrete staircase leading down to a dark tunnel entrance, flanked by metal railings on both sides. The ground surrounding the stairs is covered in grass. Two individuals are present at the top of the stairs; one is wearing a green jacket and a beanie, while the other is dressed in black clothing and a white hat. The person in the green jacket begins to descend the stairs first, followed by the individual in black. As they move down, the person in the green jacket pauses midway, looking back towards the camera, while the person in black continues their descent. The video concludes with both individuals still descending the stairs, with the person in the green jacket still pausing and looking back."
}
\end{lstlisting}

\begin{lstlisting}
Example of the intergrated annotations by Qwen-turbo:
"2018-03-15_10_131": {
        "timestamps": [
            [
                3.2,
                7.7
            ],
            [
                50.0,
                54.0
            ],
            [
                114.0,
                118.0
            ],
            [
                148.0,
                156.0
            ],
            [
                188.0,
                194.0
            ]
        ],
        "sentences": [
            "A man wearing dark clothing and a beanie hat walks down a set of concrete stairs enclosed by metal railings on either side. The stairs feature red handrails, and the man descends them steadily. As he moves downward......",
            "The video begins with a view of a concrete staircase descending into a dark tunnel. Metal railings line both sides of the staircase, and the area at the top of the stairs is overgrown with grass......",
            "The video begins with a view of a concrete staircase leading down to a dark tunnel entrance.........",
            "The video begins with a view of a concrete staircase leading down to a dark tunnel entrance. Metal railings flank both sides of the staircase......",
            "......The video ends with both individuals still moving down the stairs, with the person in the green jacket continuing to pause and look back."
        ]
    }
\end{lstlisting}

\section{Data Generation Details}~\label{AppendixB}

\begin{table}
    \centering
    \caption{Designed Prompts to guide question-answer pair generation.}
    \resizebox{0.95\textwidth}{!}{
    \begin{tabular}{p{2cm}p{4cm}p{4cm}p{3cm}}
        \toprule
        \textbf{Type} & \textbf{Task Description} & \textbf{Key Instructions} & \textbf{Example Questions} \\
        \midrule
        \multicolumn{4}{c}{\textbf{Anomaly QA}} \\
        \midrule
        Anomaly Detection &  Determine if an event related to violence, crime, or danger occurs. & Answer with 'Yes' or 'No' based on the presence of violence, crime, or danger. & Does this video contain any potentially violent or criminal activities? \\
        \midrule
        Anomaly Classification & Identify and classify any detected dangerous event using predefined categories. & Classify the anomaly into categories like Abuse, Assault, etc., or return 'None'. & What type of abnormal event is present in the video? \\
        \midrule
        Anomaly Subject &  Identify the primary subject involved in the abnormal event. & List key subjects involved in the anomaly, or return 'None' if no anomaly is detected. & Who is the main person involved in the unusual event? \\
        \midrule
        Anomaly Description &  Provide a detailed description of the detected anomaly, including setting and actions. & Describe the event, environment, and actions in detail, or return 'None'. & What is happening in the detected abnormal event? \\
        \midrule
        Anomaly Cause & Logically infer the root cause of the detected abnormal event. & Analyze environmental factors and interactions to explain the cause, or return 'None'. & What led to the unusual event occurring in the video? \\
        \midrule
        Anomaly Result &  Infer and describe the outcome of the detected abnormal event. & Describe the consequence, situation evolution, and impacts, or return 'None'. & What happens as a result of the abnormal event? \\
        \midrule
        \multicolumn{4}{c}{\textbf{Normal QA}} \\
        \midrule
        Summary QA &  Generate questions to extract a detailed description of the entire video content. & Generate three questions targeting the full sequence, with answers integrating all details. & Can you describe the entire video in detail from start to finish? \\
        \midrule
        Generic QA  & Generate questions focusing on significant aspects like appearance and motion. & Generate three questions on different aspects (appearance, motion, reasoning), with detailed answers. & Describe the entire process the person goes through from start to finish. \\
        \midrule
        Temporal QA  & Generate questions focusing on the sequence and timing of events. & Generate three questions using time references (beginning, middle, end), with answers based on the caption. & When does the main character start the primary task, and what leads up to it? \\
        \midrule
        Short Temporal QA  & Generate concise questions focusing on specific temporal events in the video. & Generate three questions on temporal aspects using approximate time references, with answers based on the caption. & When does event x happen in the video? \\
        \midrule
        Spatial QA &  Generate questions focusing on spatial details like colors and outfits. & Generate three questions on spatial aspects (colors, attire, location), with detailed answers. & What is the color of the woman's shirt? \\
        \midrule
        Reasoning QA & Generate questions focusing on actions, objects, and reasoning behind events. & Generate three questions on actions, objects, and reasoning, with concise answers including context. & Why did the player throw the ball? \\
        \bottomrule
    \end{tabular}
    }
    \label{Designed_prompts}
\end{table}

For this normal case, the generated JSON file includes the following QA categories:
\begin{itemize}

\item summary\_qa\_pairs: Cover the full narrative of the video, summarizing the entire sequence of events.
\item generic\_qa\_pairs: Focus on essential visual and behavioral information, including appearance, actions, trajectories, and inferred intentions.
\item temporal\_qa\_pairs: Address the order and timing of events, using general time references (e.g., beginning, middle, end).
\item spatial\_qa\_pairs: Explore spatial details such as clothing colors, physical positions, and scene layout.
\item reasoning\_qa\_pairs: Emphasize causal and inferential questions related to actions, locations, and motivations ("what," "where," "why").
\item short\_temporal\_qa\_pairs: Provide concise questions about specific moments or transitions within the video.
\end{itemize}
For abnormal video segments, six QA categories are defined, with one QA pair per category to ensure a concise yet focused representation of key anomalous elements. Taking the MSAD video "MSAD\_anomaly\_videos\_Shooting\_169.mp4" as an example, which depicts a shooting incident in a street setting, the scene involves multiple vehicles (a silver SUV, white SUV, and motorcycle) and several individuals (including men and women dressed in black). The video captures a violent exchange involving a firearm threat and return fire.

The corresponding JSON file for this abnormal case includes the following QA categories and content:
\begin{itemize}

\item get\_anomaly\_detection: Determines whether the video contains violent, criminal, or dangerous events.
\item get\_anomaly\_classification\_prompt: Classifies the anomaly type—in this case, as a shooting.
\item get\_anomaly\_subject: Identifies the key individuals involved in the anomaly—two men on a motorcycle (initiators with firearms) and a woman who returns fire.
\item get\_anomaly\_description: Provides a detailed account of the shooting (focused between 35.1–41.0 seconds), describing the environment (street), appearances (black clothing), and actions (threatening, shooting, fleeing).
\item get\_anomaly\_cause: Infers the likely cause of the anomaly—here, the armed threat initiated by the motorcyclists.
\item get\_anomaly\_result: Analyzes the consequence or outcome of the anomalous event.
\end{itemize}
This streamlined QA structure for abnormal segments allows the dataset to efficiently highlight critical details necessary for real-time anomaly detection and situational understanding, contrasting with the more comprehensive QA setup used for normal videos.

\section{More Data statistics}~\label{AppendixC}

\begin{figure}
\centering
\includegraphics[width=.75\textwidth]{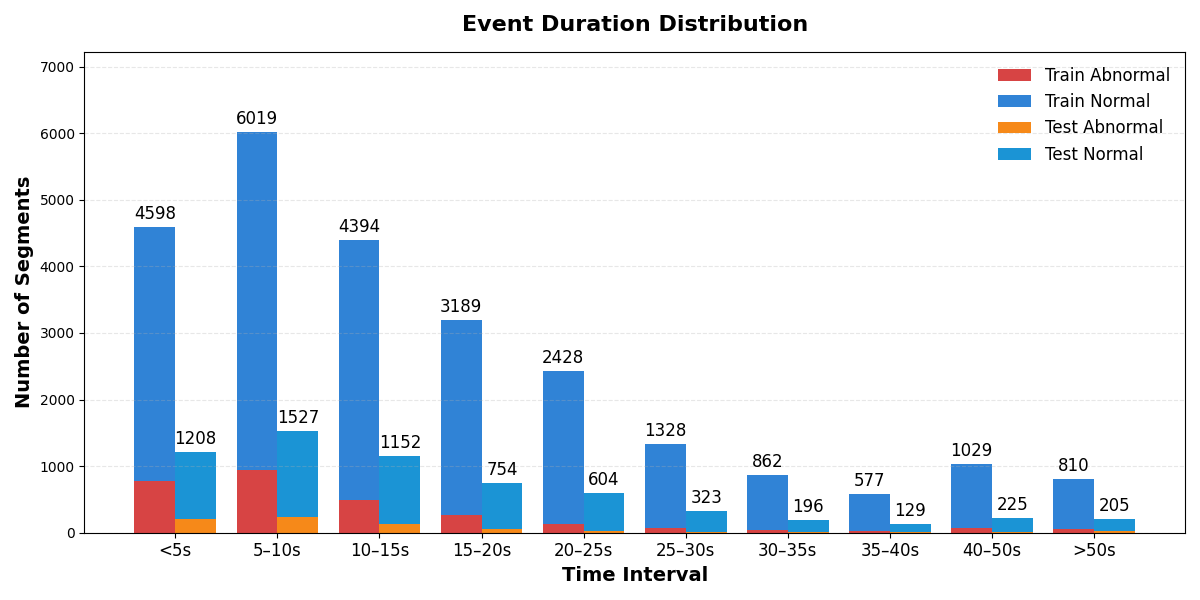}
\caption{Distribution of event durations on the training/test sets }
\label{time_bar}
\end{figure}

Figure~\ref{time_bar} illustrates the distribution of event durations on the training/test sets in this dataset. The majority of events are concentrated between 5 and 20 seconds, with a particularly high concentration in the 5-10 seconds and 10-15 seconds intervals. Notably, normal events in the training set dominate the distribution, making up the largest proportion of the dataset.

Regarding video event categorization (as shown in Table~\ref{Statistics on abnormal}), in addition to the Normal category representing non-anomalous cases, we establish a classification system comprising 18 distinct abnormal categories for the Abnormal class. These include: Abuse, Arrest, Assault, Burglary, Explosion, Fighting, Fire, Object Falling, People Falling, Pursuit, Robbery, Shooting, Shoplifting, Stealing, Traffic Accident, Threat, Vandalism, and Water Incident—covering a wide range of incidents from violent behaviors (e.g., Assault, Fighting) to environmental hazards (e.g., Fire, Water Incident).

Notably, we have observed that some abnormal categories generated by Qwen-Max exhibit redundancy or lack general applicability. To improve the accuracy of statistical analysis and standardize the taxonomy, we perform category consolidation and refinement. For instance, “Chasing” and “Chase” are unified under Pursuit due to semantic equivalence, while overly broad categories such as “Emergency Situation” and “Weapon Present” are excluded from statistical summaries due to their ambiguity in defining concrete abnormal behaviors. It is important to emphasize that this refinement is applied only during the statistical analysis phase, and the original Qwen-Max outputs remain unaltered.

\begin{table}[!ht]
    \centering
    \caption{Statistics on abnormal categories and numbers}
    \resizebox{0.5\textwidth}{!}{
    \begin{tabular}{lrrrr} 
        \toprule
        \textbf{Category} & \textbf{MEVA} & \textbf{MSAD} & \textbf{NWPU} & \textbf{UCA} \\
        \midrule
        \textbf{normal} & 2044 & 1417 & 4121 & 20384 \\
        \textbf{Abuse} & 0 & 1 & 0 & 138 \\
        \textbf{Arrest} & 0 & 2 & 0 & 143 \\
        \textbf{Assault} & 0 & 38 & 4 & 451 \\
        \textbf{Burglary} & 0 & 7 & 4 & 204 \\
        \textbf{Explosion} & 0 & 15 & 0 & 73 \\
        \textbf{Fighting} & 1 & 34 & 6 & 400 \\
        \textbf{Fire} & 0 & 52 & 0 & 217 \\
        \textbf{Object Falling} & 1 & 55 & 1 & 101 \\
        \textbf{People Falling} & 3 & 109 & 9 & 570 \\
        \textbf{Pursuit} & 0 & 1 & 1 & 19 \\
        \textbf{Robbery} & 1 & 63 & 7 & 575 \\
        \textbf{Shooting} & 0 & 15 & 0 & 56 \\
        \textbf{Shoplifting} & 0 & 0 & 0 & 26 \\
        \textbf{Stealing} & 2 & 5 & 13 & 250 \\
        \textbf{Traffic Accident} & 4 & 52 & 2 & 441 \\
        \textbf{Threat} & 0 & 1 & 0 & 7 \\
        \textbf{Vandalism} & 2 & 25 & 2 & 255 \\
        \textbf{Water Incident} & 0 & 6 & 0 & 6 \\
        \bottomrule
    \end{tabular}}
    \label{Statistics on abnormal}
\end{table}

\section{More Experiment Results}~\label{AppendixD}

More results with five metrics including CI, DO, CU, TU, AVG, across normal QA tasks normal video clips
have been shown in Table~\ref{tab:all_normal_qa_results}.

\begin{table*}[htbp]
\centering
\caption{Performance of different vision-language models across normal QA tasks on normal video clips. All values are shown in blue, with the highest value in each row bolded. $^{\dagger}$ represents our finetuned LVLMs.}
\Huge{
\resizebox{1.\textwidth}{!}{
\renewcommand\arraystretch{1.2}
\begin{tabular}{ccccc>{\columncolor{gray!30}}ccccc>{\columncolor{gray!30}}c}
\toprule
\multirow{2}{*}{\textbf{Task}} & \multirow{2}{*}{\textbf{Metric}} & 
\textbf{LLavA-OV-}  & 
\textbf{InternVL}  & 
\textbf{VideoLLa}  & 
\textbf{Qwen2.5-VL-}  & 
\textbf{LLaVA-}  & 
\textbf{LLavA-OV-}  & 
\textbf{LLaVA-} & 
\textbf{VideoLLa} & 
\textbf{Qwen2.5-VL-} \\
 & & \textbf{Qwen2-0.5B} & 
\textbf{2.5-2B} & 
\textbf{MA3-2B} & 
\textbf{3B-Instruct} & 
\textbf{Next-7B} & 
\textbf{Qwen2-7B-ov} & 
\textbf{Video-7B} & 
\textbf{MA3-7B} & 
\textbf{3B-Instruct}$^{\dagger}$ \\
\midrule
\multirow{5}{*}{\textbf{SummaryQA}} & 
CI  & {\color{normalblue}2.95} & {\color{normalblue}0.45} & {\color{normalblue}2.65} & {\color{normalblue}2.29} & {\color{normalblue}2.32} & \textbf{\color{normalblue}3.34} & {\color{normalblue}3.30} & {\color{normalblue}2.82} & {\color{normalblue}2.77}\\
 & DO  & {\color{normalblue}2.63} & {\color{normalblue}0.48} & {\color{normalblue}2.35} & {\color{normalblue}2.12} & {\color{normalblue}2.08} & \textbf{\color{normalblue}2.92} & {\color{normalblue}2.78} & {\color{normalblue}2.62} & {\color{normalblue}2.53}\\
 & CU  & {\color{normalblue}2.88} & {\color{normalblue}0.75} & {\color{normalblue}2.64} & {\color{normalblue}2.36} & {\color{normalblue}2.34} & {\color{normalblue}3.21} & \textbf{\color{normalblue}3.22} & {\color{normalblue}2.79} & {\color{normalblue}2.73}\\
 & TU  & {\color{normalblue}2.61} & {\color{normalblue}0.55} & {\color{normalblue}2.33} & {\color{normalblue}2.03} & {\color{normalblue}1.93} & \textbf{\color{normalblue}2.91} & \textbf{\color{normalblue}2.91} & {\color{normalblue}2.51} & {\color{normalblue}2.45}\\
 & Avg. & {\color{normalblue}2.76} & {\color{normalblue}0.56} & {\color{normalblue}2.49} & {\color{normalblue}2.20} & {\color{normalblue}2.17} & \textbf{\color{normalblue}3.10} & {\color{normalblue}3.05} & {\color{normalblue}2.68} & {\color{normalblue}2.62}\\
\hline
\multirow{5}{*}{\textbf{Generic QA}}&CI  & {\color{normalblue}2.97} & {\color{normalblue}2.27} & {\color{normalblue}3.02} & {\color{normalblue}2.83} & {\color{normalblue}3.11} & \textbf{\color{normalblue}3.27} & {\color{normalblue}3.19} & {\color{normalblue}3.20} & {\color{normalblue}2.92}\\
 & DO  & {\color{normalblue}2.60} & {\color{normalblue}2.13} & {\color{normalblue}2.70} & {\color{normalblue}2.56} & {\color{normalblue}2.86} & \textbf{\color{normalblue}2.91} & {\color{normalblue}2.79} & {\color{normalblue}2.87} & {\color{normalblue}2.67}\\
 & CU  & {\color{normalblue}2.90} & {\color{normalblue}2.32} & {\color{normalblue}2.94} & {\color{normalblue}2.77} & {\color{normalblue}3.03} & \textbf{\color{normalblue}3.17} & {\color{normalblue}3.10} & {\color{normalblue}3.11} & {\color{normalblue}2.85}\\
 & TU  & {\color{normalblue}2.65} & {\color{normalblue}2.06} & {\color{normalblue}2.70} & {\color{normalblue}2.50} & {\color{normalblue}2.73} & \textbf{\color{normalblue}2.94} & {\color{normalblue}2.88} & {\color{normalblue}2.87} & {\color{normalblue}2.58}\\
 & Avg. & {\color{normalblue}2.78} & {\color{normalblue}2.20} & {\color{normalblue}2.84} & {\color{normalblue}2.66} & {\color{normalblue}2.93} & \textbf{\color{normalblue}3.08} & {\color{normalblue}2.99} & {\color{normalblue}3.01} & {\color{normalblue}2.76}\\
\hline
\multirow{5}{*}{\textbf{Temporal QA}} &
CI  & {\color{normalblue}2.73} & {\color{normalblue}1.83} & {\color{normalblue}2.84} & {\color{normalblue}2.77} & {\color{normalblue}2.68} & \textbf{\color{normalblue}3.08} & {\color{normalblue}2.98} & {\color{normalblue}2.91} & {\color{normalblue}2.97}\\
 & DO  & {\color{normalblue}2.52} & {\color{normalblue}1.90} & {\color{normalblue}2.66} & {\color{normalblue}2.61} & {\color{normalblue}2.50} & \textbf{\color{normalblue}2.85} & {\color{normalblue}2.75} & {\color{normalblue}2.70} & {\color{normalblue}2.80}\\
 & CU  & {\color{normalblue}2.73} & {\color{normalblue}2.03} & {\color{normalblue}2.83} & {\color{normalblue}2.75} & {\color{normalblue}2.67} & \textbf{\color{normalblue}3.03} & {\color{normalblue}2.95} & {\color{normalblue}2.89} & {\color{normalblue}2.92}\\
 & TU  & {\color{normalblue}2.48} & {\color{normalblue}1.80} & {\color{normalblue}2.61} & {\color{normalblue}2.51} & {\color{normalblue}2.38} & \textbf{\color{normalblue}2.84} & {\color{normalblue}2.74} & {\color{normalblue}2.72} & {\color{normalblue}2.68}\\
 & Avg. & {\color{normalblue}2.62} & {\color{normalblue}1.89} & {\color{normalblue}2.73} & {\color{normalblue}2.66} & {\color{normalblue}2.56} & \textbf{\color{normalblue}2.95} & {\color{normalblue}2.85} & {\color{normalblue}2.81} & {\color{normalblue}2.84}\\
\hline
\multirow{5}{*}{\textbf{Short Temporal}}  & 
CI  & {\color{normalblue}2.71} & {\color{normalblue}2.04} & {\color{normalblue}2.73} & {\color{normalblue}2.79} & {\color{normalblue}2.70} & \textbf{\color{normalblue}2.92} & {\color{normalblue}2.90} & {\color{normalblue}2.76} & {\color{normalblue}3.04}\\
 & DO  & {\color{normalblue}2.48} & {\color{normalblue}2.00} & {\color{normalblue}2.45} & \textbf{\color{normalblue}2.66} & {\color{normalblue}2.67} & {\color{normalblue}2.63} & {\color{normalblue}2.65} & {\color{normalblue}2.46} & {\color{normalblue}2.93}\\
 & CU  & {\color{normalblue}2.76} & {\color{normalblue}2.27} & {\color{normalblue}2.74} & {\color{normalblue}2.80} & {\color{normalblue}2.78} & {\color{normalblue}2.92} & \textbf{\color{normalblue}2.94} & {\color{normalblue}2.79} & {\color{normalblue}3.00}\\
 & TU  & {\color{normalblue}2.46} & {\color{normalblue}2.00} & {\color{normalblue}2.51} & {\color{normalblue}2.52} & {\color{normalblue}2.50} & \textbf{\color{normalblue}2.68} & {\color{normalblue}2.65} & {\color{normalblue}2.56} & {\color{normalblue}2.75}\\
 & Avg. & {\color{normalblue}2.60} & {\color{normalblue}2.08} & {\color{normalblue}2.61} & {\color{normalblue}2.69} & {\color{normalblue}2.66} & \textbf{\color{normalblue}2.79} & {\color{normalblue}2.78} & {\color{normalblue}2.65} & {\color{normalblue}2.93}\\
\hline
\multirow{5}{*}{\textbf{Spatial QA}} & 
CI  & {\color{normalblue}3.22} & {\color{normalblue}1.90} & {\color{normalblue}3.07} & {\color{normalblue}2.94} & {\color{normalblue}3.03} & {\color{normalblue}3.41} & \textbf{\color{normalblue}3.60} & {\color{normalblue}3.16} & {\color{normalblue}2.97}\\
 & DO  & {\color{normalblue}2.96} & {\color{normalblue}1.83} & {\color{normalblue}2.80} & {\color{normalblue}2.78} & {\color{normalblue}2.88} & {\color{normalblue}3.17} & \textbf{\color{normalblue}3.30} & {\color{normalblue}2.91} & {\color{normalblue}2.85}\\
 & CU  & {\color{normalblue}3.23} & {\color{normalblue}2.10} & {\color{normalblue}3.10} & {\color{normalblue}2.97} & {\color{normalblue}3.06} & {\color{normalblue}3.41} & \textbf{\color{normalblue}3.57} & {\color{normalblue}3.18} & {\color{normalblue}2.98}\\
 & TU  & {\color{normalblue}3.02} & {\color{normalblue}1.85} & {\color{normalblue}2.89} & {\color{normalblue}2.74} & {\color{normalblue}2.83} & {\color{normalblue}3.23} & \textbf{\color{normalblue}3.40} & {\color{normalblue}2.98} & {\color{normalblue}2.76}\\
 & Avg. & {\color{normalblue}3.10} & {\color{normalblue}1.92} & {\color{normalblue}2.97} & {\color{normalblue}2.86} & {\color{normalblue}2.95} & {\color{normalblue}3.31} & \textbf{\color{normalblue}3.47} & {\color{normalblue}3.06} & {\color{normalblue}2.89}\\
\hline
\multirow{5}{*}{\textbf{Reasoning QA}}  & 
CI  & {\color{normalblue}3.05} & {\color{normalblue}2.45} & {\color{normalblue}3.02} & {\color{normalblue}3.11} & {\color{normalblue}3.19} & {\color{normalblue}3.22} & \textbf{\color{normalblue}3.39} & {\color{normalblue}3.13} & {\color{normalblue}3.14}\\
 & DO  & {\color{normalblue}2.75} & {\color{normalblue}2.29} & {\color{normalblue}2.69} & {\color{normalblue}2.94} & {\color{normalblue}2.97} & {\color{normalblue}2.87} & \textbf{\color{normalblue}3.05} & {\color{normalblue}2.80} & {\color{normalblue}3.06}\\
 & CU  & {\color{normalblue}3.07} & {\color{normalblue}2.57} & {\color{normalblue}3.05} & {\color{normalblue}3.13} & {\color{normalblue}3.19} & {\color{normalblue}3.23} & \textbf{\color{normalblue}3.38} & {\color{normalblue}3.15} & {\color{normalblue}3.14}\\
 & TU  & {\color{normalblue}2.84} & {\color{normalblue}2.32} & {\color{normalblue}2.81} & {\color{normalblue}2.85} & {\color{normalblue}2.90} & {\color{normalblue}3.00} & \textbf{\color{normalblue}3.18} & {\color{normalblue}2.91} & {\color{normalblue}2.90}\\
 & Avg. & {\color{normalblue}2.93} & {\color{normalblue}2.41} & {\color{normalblue}2.89} & {\color{normalblue}3.01} & {\color{normalblue}3.06} & {\color{normalblue}3.08} & \textbf{\color{normalblue}3.25} & {\color{normalblue}2.99} & {\color{normalblue}3.06}\\
\hline
\end{tabular}
}
}
\label{tab:all_normal_qa_results}
\end{table*}

More results with five metrics including CI, DO, CU, TU, AVG, across normal QA tasks on abnormal video clips 
have been shown in Table~\ref{tab:normal_qa_results}.

\begin{table}[]
\centering
\caption{Performance of different vision-language models across normal QA tasks on abnormal video clips. All values are shown in green, with the highest value in each row bolded. $^{\dagger}$ represents our finetuned LVLMs.}
\Huge{
\resizebox{1.\textwidth}{!}{
\renewcommand\arraystretch{1.2}
\begin{tabular}{ccccc>{\columncolor{gray!30}}ccccc>{\columncolor{gray!30}}c}
\toprule
\multirow{2}{*}{\textbf{Task}} & \multirow{2}{*}{\textbf{Metric}} & 
\textbf{LLavA-OV-}  & 
\textbf{InternVL}  & 
\textbf{VideoLLa}  & 
\textbf{Qwen2.5-VL-}  & 
\textbf{LLaVA-}  & 
\textbf{LLavA-OV-}  & 
\textbf{LLaVA-} & 
\textbf{VideoLLa} & 
\textbf{Qwen2.5-VL-} \\
 & & \textbf{Qwen2-0.5B} & 
\textbf{2.5-2B} & 
\textbf{MA3-2B} & 
\textbf{3B-Instruct} & 
\textbf{Next-7B} & 
\textbf{Qwen2-7B-ov} & 
\textbf{Video-7B} & 
\textbf{MA3-7B} & 
\textbf{3B-Instruct}$^{\dagger}$ \\
\midrule
\multirow{5}{*}{Summary QA} 
& CI  & {\color{abnormalgreen}2.56} & {\color{abnormalgreen}0.22} & {\color{abnormalgreen}2.06} & {\color{abnormalgreen}1.46} & {\color{abnormalgreen}1.78} & {\color{abnormalgreen}2.97} & \textbf{\color{abnormalgreen}3.03} & {\color{abnormalgreen}2.35} & {\color{abnormalgreen}2.05}\\
& DO & {\color{abnormalgreen}2.35} & {\color{abnormalgreen}0.26} & {\color{abnormalgreen}1.92} & {\color{abnormalgreen}1.46} & {\color{abnormalgreen}1.68} & \textbf{\color{abnormalgreen}2.69} & {\color{abnormalgreen}2.63} & {\color{abnormalgreen}2.28} & {\color{abnormalgreen}1.96}\\
& CU & {\color{abnormalgreen}2.56} & {\color{abnormalgreen}0.60} & {\color{abnormalgreen}2.17} & {\color{abnormalgreen}1.68} & {\color{abnormalgreen}1.90} & {\color{abnormalgreen}2.90} & \textbf{\color{abnormalgreen}3.02} & {\color{abnormalgreen}2.43} & {\color{abnormalgreen}2.14}\\
& TU & {\color{abnormalgreen}2.24} & {\color{abnormalgreen}0.40} & {\color{abnormalgreen}1.84} & {\color{abnormalgreen}1.38} & {\color{abnormalgreen}1.52} & {\color{abnormalgreen}2.61} & \textbf{\color{abnormalgreen}2.72} & {\color{abnormalgreen}2.11} & {\color{abnormalgreen}1.82}\\
& Avg. & {\color{abnormalgreen}2.43} & {\color{abnormalgreen}0.37} & {\color{abnormalgreen}2.00} & {\color{abnormalgreen}1.49} & {\color{abnormalgreen}1.72} & {\color{abnormalgreen}2.79} & \textbf{\color{abnormalgreen}2.85} & {\color{abnormalgreen}2.29} & {\color{abnormalgreen}1.99}\\
\hline
\multirow{5}{*}{Generic QA} 
& CI  & {\color{abnormalgreen}2.66} & {\color{abnormalgreen}1.95} & {\color{abnormalgreen}2.68} & {\color{abnormalgreen}2.23} & {\color{abnormalgreen}2.72} & \textbf{\color{abnormalgreen}3.01} & {\color{abnormalgreen}2.96} & {\color{abnormalgreen}2.83} & {\color{abnormalgreen}2.43}\\
& DO & {\color{abnormalgreen}2.41} & {\color{abnormalgreen}1.90} & {\color{abnormalgreen}2.48} & {\color{abnormalgreen}2.15} & {\color{abnormalgreen}2.58} & \textbf{\color{abnormalgreen}2.74} & {\color{abnormalgreen}2.63} & {\color{abnormalgreen}2.63} & {\color{abnormalgreen}2.35}\\
& CU & {\color{abnormalgreen}2.65} & {\color{abnormalgreen}2.08} & {\color{abnormalgreen}2.67} & {\color{abnormalgreen}2.29} & {\color{abnormalgreen}2.71} & \textbf{\color{abnormalgreen}2.97} & {\color{abnormalgreen}2.94} & {\color{abnormalgreen}2.81} & {\color{abnormalgreen}2.47}\\
& TU & {\color{abnormalgreen}2.37} & {\color{abnormalgreen}1.80} & {\color{abnormalgreen}2.39} & {\color{abnormalgreen}2.01} & {\color{abnormalgreen}2.39} & \textbf{\color{abnormalgreen}2.71} & {\color{abnormalgreen}2.68} & {\color{abnormalgreen}2.53} & {\color{abnormalgreen}2.18}\\
& Avg. & {\color{abnormalgreen}2.52} & {\color{abnormalgreen}1.93} & {\color{abnormalgreen}2.55} & {\color{abnormalgreen}2.17} & {\color{abnormalgreen}2.60} & \textbf{\color{abnormalgreen}2.86} & {\color{abnormalgreen}2.80} & {\color{abnormalgreen}2.70} & {\color{abnormalgreen}2.36}\\
\hline
\multirow{5}{*}{Temporal QA} 
& CI  & {\color{abnormalgreen}2.50} & {\color{abnormalgreen}1.57} & {\color{abnormalgreen}2.53} & {\color{abnormalgreen}2.26} & {\color{abnormalgreen}2.31} & \textbf{\color{abnormalgreen}2.89} & {\color{abnormalgreen}2.86} & {\color{abnormalgreen}2.61} & {\color{abnormalgreen}2.51}\\
& DO & {\color{abnormalgreen}2.38} & {\color{abnormalgreen}1.69} & {\color{abnormalgreen}2.47} & {\color{abnormalgreen}2.27} & {\color{abnormalgreen}2.29} & \textbf{\color{abnormalgreen}2.74} & {\color{abnormalgreen}2.68} & {\color{abnormalgreen}2.50} & {\color{abnormalgreen}2.52}\\
& CU & {\color{abnormalgreen}2.57} & {\color{abnormalgreen}1.84} & {\color{abnormalgreen}2.59} & {\color{abnormalgreen}2.37} & {\color{abnormalgreen}2.40} & \textbf{\color{abnormalgreen}2.90} & {\color{abnormalgreen}2.88} & {\color{abnormalgreen}2.65} & {\color{abnormalgreen}2.57}\\
& TU & {\color{abnormalgreen}2.29} & {\color{abnormalgreen}1.60} & {\color{abnormalgreen}2.34} & {\color{abnormalgreen}2.12} & {\color{abnormalgreen}2.10} & \textbf{\color{abnormalgreen}2.69} & {\color{abnormalgreen}2.65} & {\color{abnormalgreen}2.44} & {\color{abnormalgreen}2.33}\\
& Avg. & {\color{abnormalgreen}2.43} & {\color{abnormalgreen}1.68} & {\color{abnormalgreen}2.49} & {\color{abnormalgreen}2.26} & {\color{abnormalgreen}2.28} & \textbf{\color{abnormalgreen}2.81} & {\color{abnormalgreen}2.77} & {\color{abnormalgreen}2.55} & {\color{abnormalgreen}2.48}\\
\hline
\multirow{5}{*}{Short Temporal} 
& CI  & {\color{abnormalgreen}2.51} & {\color{abnormalgreen}1.89} & {\color{abnormalgreen}2.52} & {\color{abnormalgreen}2.33} & {\color{abnormalgreen}2.48} & {\color{abnormalgreen}2.75} & \textbf{\color{abnormalgreen}2.77} & {\color{abnormalgreen}2.53} & {\color{abnormalgreen}2.61}\\
& DO & {\color{abnormalgreen}2.36} & {\color{abnormalgreen}1.89} & {\color{abnormalgreen}2.30} & {\color{abnormalgreen}2.32} & \textbf{\color{abnormalgreen}2.51} & {\color{abnormalgreen}2.52} & {\color{abnormalgreen}2.60} & {\color{abnormalgreen}2.30} & {\color{abnormalgreen}2.65}\\
& CU & {\color{abnormalgreen}2.61} & {\color{abnormalgreen}2.15} & {\color{abnormalgreen}2.60} & {\color{abnormalgreen}2.44} & {\color{abnormalgreen}2.63} & {\color{abnormalgreen}2.82} & \textbf{\color{abnormalgreen}2.87} & {\color{abnormalgreen}2.61} & {\color{abnormalgreen}2.70}\\
& TU & {\color{abnormalgreen}2.31} & {\color{abnormalgreen}1.88} & {\color{abnormalgreen}2.35} & {\color{abnormalgreen}2.15} & {\color{abnormalgreen}2.30} & {\color{abnormalgreen}2.53} & \textbf{\color{abnormalgreen}2.60} & {\color{abnormalgreen}2.37} & {\color{abnormalgreen}2.40}\\
& Avg. & {\color{abnormalgreen}2.45} & {\color{abnormalgreen}1.95} & {\color{abnormalgreen}2.44} & {\color{abnormalgreen}2.31} & {\color{abnormalgreen}2.48} & {\color{abnormalgreen}2.66} & \textbf{\color{abnormalgreen}2.71} & {\color{abnormalgreen}2.45} & {\color{abnormalgreen}2.59}\\
\hline
\multirow{5}{*}{Spatial QA} 
& CI  & {\color{abnormalgreen}3.23} & {\color{abnormalgreen}1.85} & {\color{abnormalgreen}3.09} & {\color{abnormalgreen}2.83} & {\color{abnormalgreen}3.03} & {\color{abnormalgreen}3.45} & \textbf{\color{abnormalgreen}3.65} & {\color{abnormalgreen}3.17} & {\color{abnormalgreen}3.00}\\
& DO & {\color{abnormalgreen}2.98} & {\color{abnormalgreen}1.79} & {\color{abnormalgreen}2.83} & {\color{abnormalgreen}2.68} & {\color{abnormalgreen}2.89} & {\color{abnormalgreen}3.20} & \textbf{\color{abnormalgreen}3.33} & {\color{abnormalgreen}2.91} & {\color{abnormalgreen}2.87}\\
& CU & {\color{abnormalgreen}3.23} & {\color{abnormalgreen}2.09} & {\color{abnormalgreen}3.11} & {\color{abnormalgreen}2.86} & {\color{abnormalgreen}3.06} & {\color{abnormalgreen}3.43} & \textbf{\color{abnormalgreen}3.62} & {\color{abnormalgreen}3.19} & {\color{abnormalgreen}2.97}\\
& TU & {\color{abnormalgreen}3.04} & {\color{abnormalgreen}1.82} & {\color{abnormalgreen}2.92} & {\color{abnormalgreen}2.64} & {\color{abnormalgreen}2.83} & {\color{abnormalgreen}3.27} & \textbf{\color{abnormalgreen}3.45} & {\color{abnormalgreen}2.99} & {\color{abnormalgreen}2.76}\\
& Avg. & {\color{abnormalgreen}3.12} & {\color{abnormalgreen}1.89} & {\color{abnormalgreen}2.99} & {\color{abnormalgreen}2.75} & {\color{abnormalgreen}2.95} & {\color{abnormalgreen}3.34} & \textbf{\color{abnormalgreen}3.51} & {\color{abnormalgreen}3.06} & {\color{abnormalgreen}2.89}\\
\hline
\multirow{5}{*}{Reasoning QA} 
& CI  & {\color{abnormalgreen}2.79} & {\color{abnormalgreen}2.26} & {\color{abnormalgreen}2.74} & {\color{abnormalgreen}2.75} & {\color{abnormalgreen}2.92} & {\color{abnormalgreen}3.00} & \textbf{\color{abnormalgreen}3.15} & {\color{abnormalgreen}2.86} & {\color{abnormalgreen}2.87}\\
& DO & {\color{abnormalgreen}2.53} & {\color{abnormalgreen}2.16} & {\color{abnormalgreen}2.48} & {\color{abnormalgreen}2.65} & {\color{abnormalgreen}2.72} & {\color{abnormalgreen}2.70} & \textbf{\color{abnormalgreen}2.87} & {\color{abnormalgreen}2.60} & {\color{abnormalgreen}2.86}\\
& CU & {\color{abnormalgreen}2.83} & {\color{abnormalgreen}2.43} & {\color{abnormalgreen}2.84} & {\color{abnormalgreen}2.84} & {\color{abnormalgreen}2.95} & {\color{abnormalgreen}3.04} & \textbf{\color{abnormalgreen}3.21} & {\color{abnormalgreen}2.95} & {\color{abnormalgreen}3.00}\\
& TU & {\color{abnormalgreen}2.60} & {\color{abnormalgreen}2.17} & {\color{abnormalgreen}2.57} & {\color{abnormalgreen}2.57} & {\color{abnormalgreen}2.64} & {\color{abnormalgreen}2.81} & \textbf{\color{abnormalgreen}2.98} & {\color{abnormalgreen}2.70} & {\color{abnormalgreen}2.72}\\
& Avg. & {\color{abnormalgreen}2.69} & {\color{abnormalgreen}2.26} & {\color{abnormalgreen}2.66} & {\color{abnormalgreen}2.70} & {\color{abnormalgreen}2.81} & {\color{abnormalgreen}2.89} & \textbf{\color{abnormalgreen}3.05} & {\color{abnormalgreen}2.78} & {\color{abnormalgreen}2.85}\\
\hline
\end{tabular}}}
\label{tab:normal_qa_results}
\end{table}

More results with five metrics including CI, DO, CU, TU, AVG, across Normal QA tasks on abnormal video clips 
have been shown in Table~\ref{tab:abnormal_qa_pairs}.

\begin{table}
\centering
\caption{Performance of different vision-language models across abnormal QA tasks on abnormal video clips. All values are shown in brown, with the highest value in each row bolded. $^{\dagger}$ represents our finetuned LVLMs.}
\Huge{
\resizebox{1.\textwidth}{!}{
\renewcommand\arraystretch{1.2}
\begin{tabular}{ccccc>{\columncolor{gray!30}}ccccc>{\columncolor{gray!30}}c}
\toprule
\multirow{2}{*}{\textbf{Task}} & \multirow{2}{*}{\textbf{Metric}} & 
\textbf{LLavA-OV-}  & 
\textbf{InternVL}  & 
\textbf{VideoLLa}  & 
\textbf{Qwen2.5-VL-}  & 
\textbf{LLaVA-}  & 
\textbf{LLavA-OV-}  & 
\textbf{LLaVA-} & 
\textbf{VideoLLa} & 
\textbf{Qwen2.5-VL-} \\
 & & \textbf{Qwen2-0.5B} & 
\textbf{2.5-2B} & 
\textbf{MA3-2B} & 
\textbf{3B-Instruct} & 
\textbf{Next-7B} & 
\textbf{Qwen2-7B-ov} & 
\textbf{Video-7B} & 
\textbf{MA3-7B} & 
\textbf{3B-Instruct}$^{\dagger}$ \\
\midrule
\multirow{5}{*}{Detection QA} 
& CI  & \textbf{\color{browncolor}3.10} & {\color{browncolor}1.97} & {\color{browncolor}1.90} & {\color{browncolor}1.92} & {\color{browncolor}2.39} & {\color{browncolor}2.54} & {\color{browncolor}1.95} & {\color{browncolor}2.07} &  {\color{browncolor}1.57}\\
& DO & {\color{browncolor}2.69} & {\color{browncolor}1.71} & {\color{browncolor}1.81} & {\color{browncolor}1.93} & \textbf{\color{browncolor}2.51} & {\color{browncolor}2.50} & {\color{browncolor}1.85} & {\color{browncolor}1.97} & {\color{browncolor}1.75}\\
& CU & \textbf{\color{browncolor}3.11} & {\color{browncolor}2.00} & {\color{browncolor}1.94} & {\color{browncolor}1.88} & {\color{browncolor}2.29} & {\color{browncolor}2.54} & {\color{browncolor}1.96} & {\color{browncolor}2.11} & {\color{browncolor}1.54}\\
& TU & \textbf{\color{browncolor}2.86} & {\color{browncolor}1.83} & {\color{browncolor}1.81} & {\color{browncolor}1.69} & {\color{browncolor}2.09} & {\color{browncolor}2.52} & {\color{browncolor}1.90} & {\color{browncolor}1.95} & {\color{browncolor}1.45}\\
& Avg. & \textbf{\color{browncolor}2.94} & {\color{browncolor}1.88} & {\color{browncolor}1.87} & {\color{browncolor}1.85} & {\color{browncolor}2.32} & {\color{browncolor}2.53} & {\color{browncolor}1.92} & {\color{browncolor}2.03} & {\color{browncolor}1.58}\\
\midrule
\multirow{5}{*}{Classification QA} 
& CI  & {\color{browncolor}2.36} & {\color{browncolor}0.86} & {\color{browncolor}1.88} & {\color{browncolor}2.00} & {\color{browncolor}2.47} & \textbf{\color{browncolor}2.52} & {\color{browncolor}2.47} & {\color{browncolor}1.37} & {\color{browncolor}1.90}\\
& DO & {\color{browncolor}2.53} & {\color{browncolor}1.22} & {\color{browncolor}2.07} & {\color{browncolor}2.40} & {\color{browncolor}2.74} & {\color{browncolor}2.61} & \textbf{\color{browncolor}2.72} & {\color{browncolor}1.49} & {\color{browncolor}2.42}\\
& CU & {\color{browncolor}2.55} & {\color{browncolor}1.36} & {\color{browncolor}2.22} & {\color{browncolor}2.30} & {\color{browncolor}2.63} & \textbf{\color{browncolor}2.77} & {\color{browncolor}2.71} & {\color{browncolor}1.84} & {\color{browncolor}2,18}\\
& TU & {\color{browncolor}2.43} & {\color{browncolor}1.05} & {\color{browncolor}1.94} & {\color{browncolor}2.05} & {\color{browncolor}2.41} & \textbf{\color{browncolor}2.48} & {\color{browncolor}2.45} & {\color{browncolor}1.41} & {\color{browncolor}2.01}\\
& Avg. & {\color{browncolor}2.47} & {\color{browncolor}1.12} & {\color{browncolor}2.03} & {\color{browncolor}2.19} & {\color{browncolor}2.56} & \textbf{\color{browncolor}2.60} & {\color{browncolor}2.59} & {\color{browncolor}1.53} & {\color{browncolor}2.13}\\
\midrule
\multirow{5}{*}{Subject QA} 
& CI  & {\color{browncolor}2.76} & {\color{browncolor}1.30} & {\color{browncolor}2.49} & {\color{browncolor}2.24} & {\color{browncolor}2.69} & {\color{browncolor}2.75} & \textbf{\color{browncolor}2.83} & {\color{browncolor}2.71} & {\color{browncolor}2.55}\\
& DO & {\color{browncolor}2.52} & {\color{browncolor}1.46} & {\color{browncolor}2.32} & {\color{browncolor}2.23} & {\color{browncolor}2.56} & {\color{browncolor}2.51} & \textbf{\color{browncolor}2.59} & {\color{browncolor}2.53} & {\color{browncolor}2.69}\\
& CU & {\color{browncolor}2.87} & {\color{browncolor}1.74} & {\color{browncolor}2.64} & {\color{browncolor}2.43} & {\color{browncolor}2.75} & {\color{browncolor}2.86} & \textbf{\color{browncolor}2.93} & {\color{browncolor}2.83} & {\color{browncolor}2.66}\\
& TU & {\color{browncolor}2.62} & {\color{browncolor}1.36} & {\color{browncolor}2.38} & {\color{browncolor}2.07} & {\color{browncolor}2.36} & {\color{browncolor}2.62} & \textbf{\color{browncolor}2.71} & {\color{browncolor}2.55} & {\color{browncolor}2.35}\\
& Avg. & {\color{browncolor}2.69} & {\color{browncolor}1.46} & {\color{browncolor}2.46} & {\color{browncolor}2.24} & {\color{browncolor}2.59} & {\color{browncolor}2.68} & \textbf{\color{browncolor}2.76} & {\color{browncolor}2.66} & {\color{browncolor}2.56}\\
\midrule
\multirow{5}{*}{Description QA} 
& CI  & {\color{browncolor}2.56} & {\color{browncolor}0.35} & {\color{browncolor}1.98} & {\color{browncolor}1.76} & {\color{browncolor}2.21} & {\color{browncolor}2.71} & \textbf{\color{browncolor}2.92} & {\color{browncolor}2.05} & {\color{browncolor}2.11}\\
& DO & {\color{browncolor}2.28} & {\color{browncolor}0.46} & {\color{browncolor}1.73} & {\color{browncolor}1.68} & {\color{browncolor}2.06} & {\color{browncolor}2.40} & \textbf{\color{browncolor}2.58} & {\color{browncolor}1.90} & {\color{browncolor}2.03}\\
& CU & {\color{browncolor}2.61} & {\color{browncolor}0.81} & {\color{browncolor}2.10} & {\color{browncolor}1.88} & {\color{browncolor}2.27} & {\color{browncolor}2.67} & \textbf{\color{browncolor}2.92} & {\color{browncolor}2.23} & {\color{browncolor}2.17}\\
& TU & {\color{browncolor}2.27} & {\color{browncolor}0.55} & {\color{browncolor}1.76} & {\color{browncolor}1.64} & {\color{browncolor}1.92} & {\color{browncolor}2.39} & \textbf{\color{browncolor}2.62} & {\color{browncolor}1.88} & {\color{browncolor}1.91}\\
& Avg. & {\color{browncolor}2.43} & {\color{browncolor}0.55} & {\color{browncolor}1.89} & {\color{browncolor}1.74} & {\color{browncolor}2.11} & {\color{browncolor}2.54} & \textbf{\color{browncolor}2.76} & {\color{browncolor}2.01} & {\color{browncolor}2.06}\\
\midrule
\multirow{5}{*}{Cause QA} 
& CI  & {\color{browncolor}1.69} & {\color{browncolor}0.46} & {\color{browncolor}1.31} & {\color{browncolor}1.29} & \textbf{\color{browncolor}2.03} & {\color{browncolor}1.51} & {\color{browncolor}1.56} & {\color{browncolor}1.69} & {\color{browncolor}1.64}\\
& DO & {\color{browncolor}1.49} & {\color{browncolor}0.55} & {\color{browncolor}1.23} & {\color{browncolor}1.28} & \textbf{\color{browncolor}1.87} & {\color{browncolor}1.16} & {\color{browncolor}1.24} & {\color{browncolor}1.71} & {\color{browncolor}1.69}\\
& CU & {\color{browncolor}1.87} & {\color{browncolor}0.87} & {\color{browncolor}1.55} & {\color{browncolor}1.40} & \textbf{\color{browncolor}2.06} & {\color{browncolor}1.66} & {\color{browncolor}1.72} & {\color{browncolor}1.86} & {\color{browncolor}1.79}\\
& TU & {\color{browncolor}1.62} & {\color{browncolor}0.69} & {\color{browncolor}1.38} & {\color{browncolor}1.29} & \textbf{\color{browncolor}1.86} & {\color{browncolor}1.51} & {\color{browncolor}1.54} & {\color{browncolor}1.71} & {\color{browncolor}1.63}\\
& Avg. & {\color{browncolor}1.67} & {\color{browncolor}0.64} & {\color{browncolor}1.37} & {\color{browncolor}1.32} & \textbf{\color{browncolor}1.96} & {\color{browncolor}1.46} & {\color{browncolor}1.52} & {\color{browncolor}1.75} & {\color{browncolor}1.69}\\
\midrule
\multirow{5}{*}{Result QA} 
& CI  & {\color{browncolor}1.62} & {\color{browncolor}0.53} & {\color{browncolor}1.20} & {\color{browncolor}1.08} & {\color{browncolor}1.73} & {\color{browncolor}1.63} & \textbf{\color{browncolor}1.89} & {\color{browncolor}1.25} & {\color{browncolor}1.25}\\
& DO & {\color{browncolor}1.39} & {\color{browncolor}0.63} & {\color{browncolor}0.91} & {\color{browncolor}1.03} & {\color{browncolor}1.49} & {\color{browncolor}1.22} & \textbf{\color{browncolor}1.56} & {\color{browncolor}0.99} & {\color{browncolor}1.19}\\
& CU & {\color{browncolor}1.84} & {\color{browncolor}0.99} & {\color{browncolor}1.47} & {\color{browncolor}1.35} & {\color{browncolor}1.87} & {\color{browncolor}1.85} & \textbf{\color{browncolor}2.07} & {\color{browncolor}1.56} & {\color{browncolor}1.53}\\
& TU & {\color{browncolor}1.54} & {\color{browncolor}0.80} & {\color{browncolor}1.15} & {\color{browncolor}1.09} & {\color{browncolor}1.51} & {\color{browncolor}1.51} & \textbf{\color{browncolor}1.74} & {\color{browncolor}1.26} & {\color{browncolor}1.24}\\
& Avg. & {\color{browncolor}1.60} & {\color{browncolor}0.74} & {\color{browncolor}1.18} & {\color{browncolor}1.13} & {\color{browncolor}1.65} & {\color{browncolor}1.55} & \textbf{\color{browncolor}1.81} & {\color{browncolor}1.27} & {\color{browncolor}1.30}\\
\bottomrule
\end{tabular}
}
}
\label{tab:abnormal_qa_pairs}
\end{table}

\begin{figure}
    \centering
    \includegraphics[width=1.\linewidth]{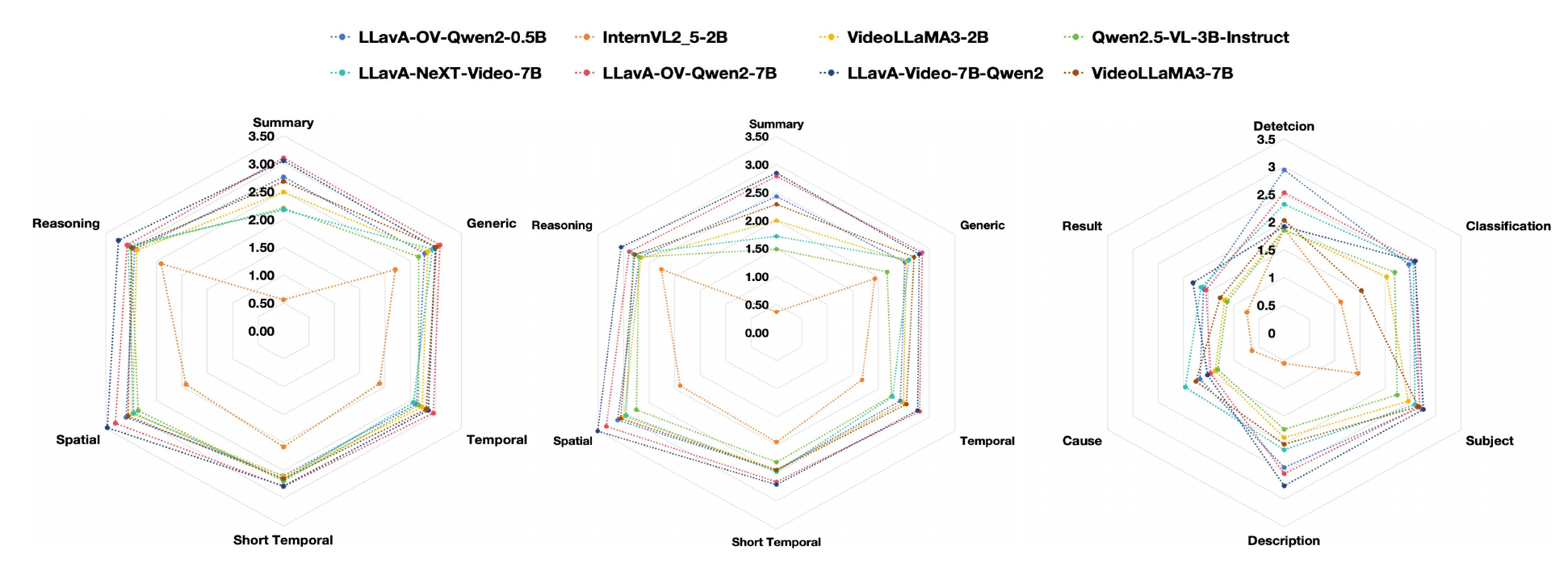}
    \caption{Comparisons of scores from different LVLMs across different QA tasks. Left: normal video clips vs. normal QA tasks. Middle: abnormal video clips vs. normal QA tasks. Right: abnormal video clips vs. abnormal QA tasks.}
    \label{fig:5}
\end{figure}

Figure~\ref{fig:5} shows the direct score comparisons of different LVLMs across different QA tasks.

\begin{figure}
    \centering
    \includegraphics[width=1.\linewidth]{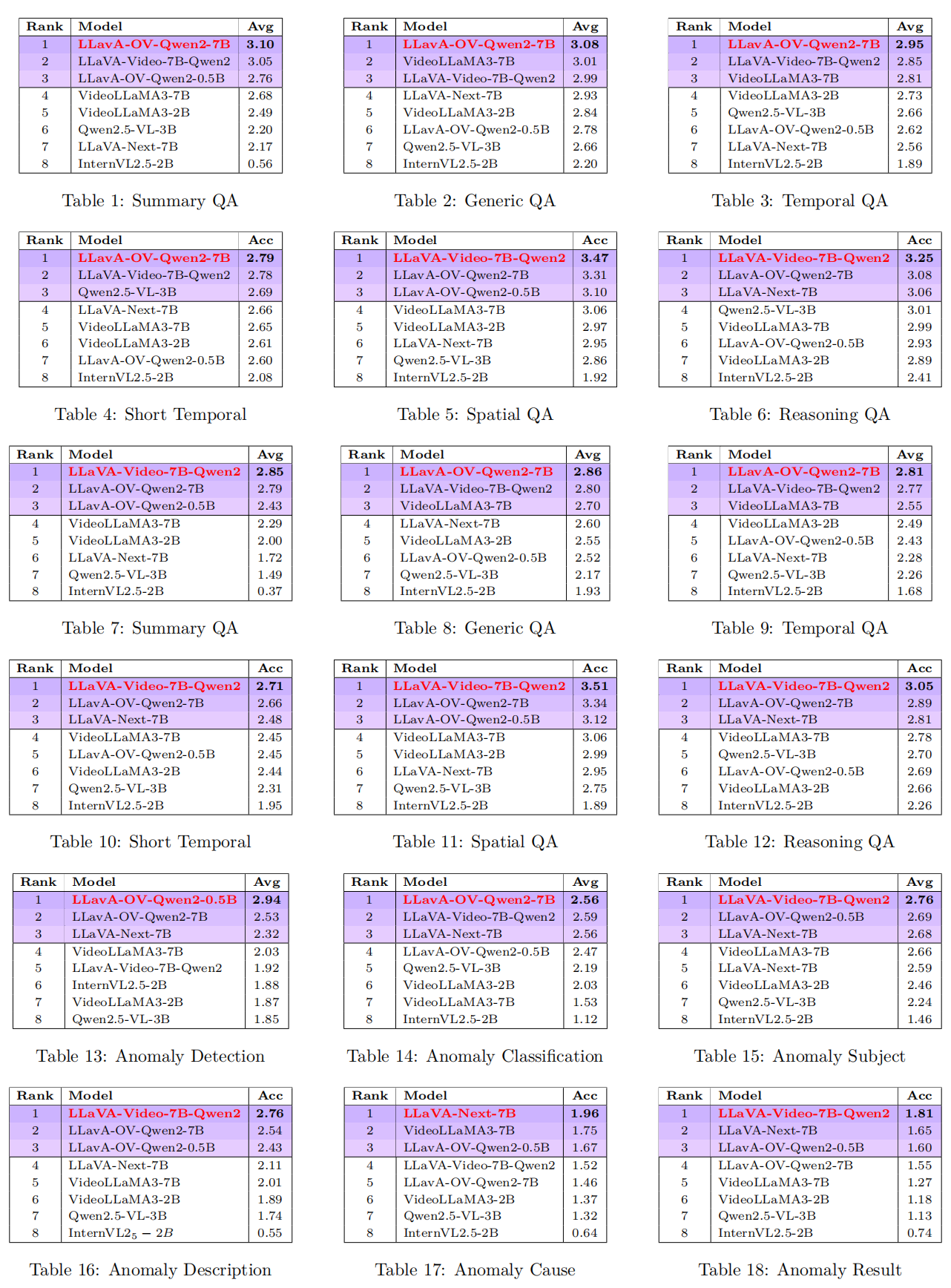}
    \caption{Performance rankings of LVLMs across different QA tasks. No.1-6 lists denote normal video clips vs. normal QA tasks. No.7-12 lists denote abnormal video clips vs. normal QA tasks. No.13-18 lists denote abnormal video clips vs. abnormal QA tasks. }
    \label{fig:6}
\end{figure}

Figure~\ref{fig:6} shows an overall performance rank of LVLMs across different QA tasks.

\section{Visualization for inference results of LVLMs.} ~\label{AppendixE}

\begin{figure}
    \centering
    \includegraphics[width=1.\linewidth]{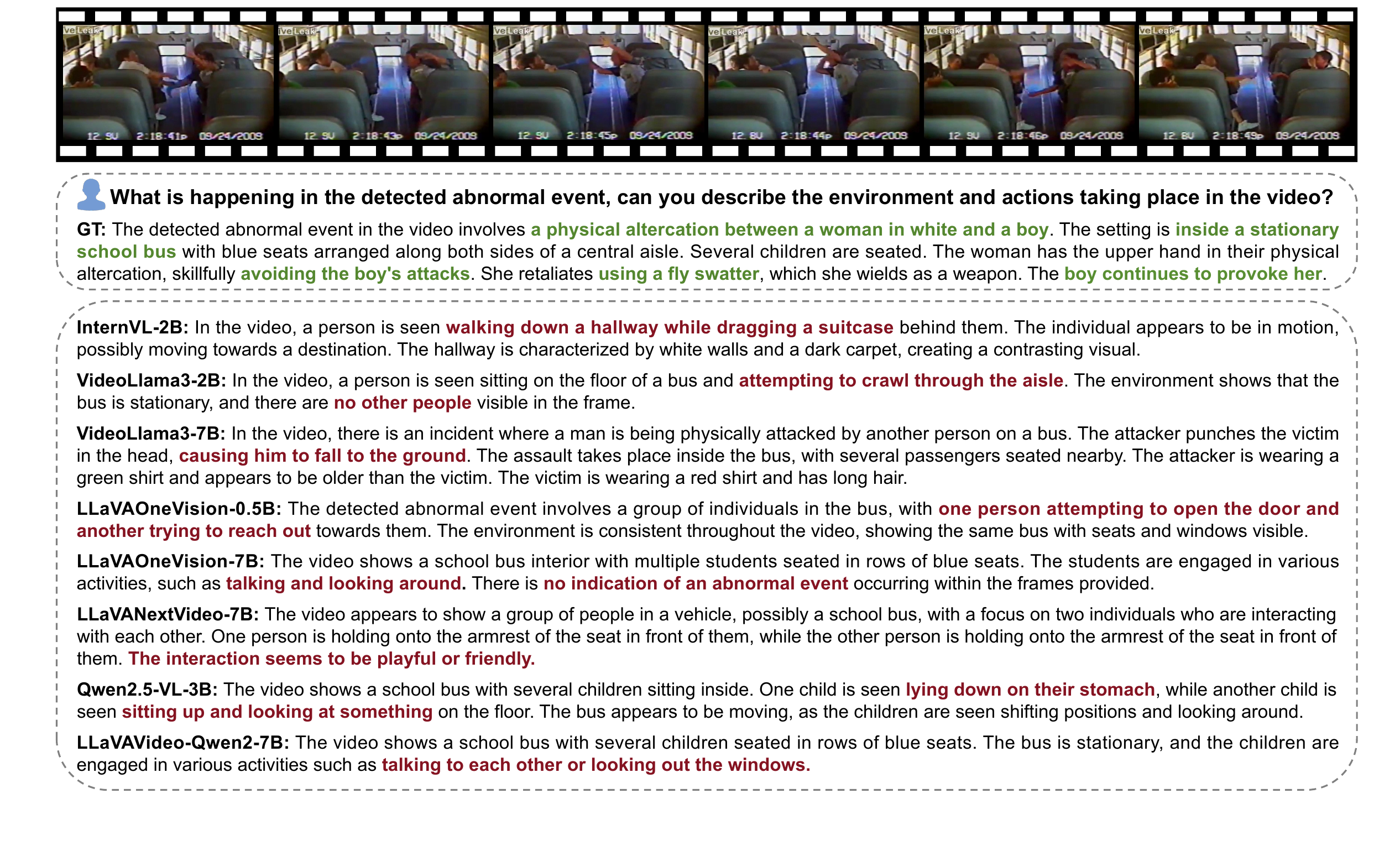}
    \caption{Abnormal videos QA Examples. All of the LVLMs provide wrong answers. }
    \label{fig:7}
\end{figure}

Figure~\ref{fig:7} shows an example of visualization for inference results of LVLMs. All of the LVLMs provide wrong answers for abnormal QA tasks. 

\clearpage
\newpage
\section{Limitations}~\label{AppendixF}

However, this study also has some limitations. Model evaluations are based on a limited dataset, and performance may vary when applied to different surveillance environments or low-quality video data. Although larger models perform well in terms of accuracy, they have high computational demands, which may make them challenging to deploy in resource-constrained environments. Future research should focus on optimizing the temporal reasoning capabilities of smaller models and further explore domain-specific fine-tuning techniques to improve performance in various monitoring scenarios.
\clearpage
\newpage

\end{document}